\pdfoutput=1
\documentclass[11pt]{article}

\usepackage[table,dvipsnames]{xcolor}
\usepackage[preprint]{acl}

\usepackage{times}
\usepackage{latexsym}

\usepackage[T1]{fontenc}
\usepackage[utf8]{inputenc}

\usepackage{microtype}

\usepackage{inconsolata}

\usepackage{graphicx}

\usepackage{amsmath}
\usepackage{amsthm}
\usepackage{amsfonts}
\usepackage{booktabs}
\usepackage{multirow}
\usepackage{xspace}
\usepackage{xcolor}
\usepackage{tcolorbox}
\usepackage{xargs}
\usepackage{MnSymbol,bbding,pifont}
\usepackage{colortbl}
\usepackage{subcaption}
\usepackage{graphicx}
\usepackage{tikz}
\usepackage{float}

\usepackage[utf8]{inputenc} % allow utf-8 input
\usepackage[T1]{fontenc}    % use 8-bit T1 fonts
\usepackage{hyperref}       % hyperlinks
\usepackage{url}            % simple URL typesetting
\usepackage{booktabs}       % professional-quality tables
\usepackage{amsfonts}       % blackboard math symbols
\usepackage{amsthm}
\usepackage{amsmath}
\usepackage{nicefrac}       % compact symbols for 1/2, etc.
\usepackage{microtype}      % microtypography
\usepackage{txfonts}
\usepackage{url}
\usepackage{booktabs}
\usepackage{changepage}
\usepackage{xargs}
\usepackage{wrapfig,lipsum,booktabs}
\usepackage{longtable}
\usepackage{subcaption}
\usepackage{tikz}
\usepackage{xspace}
\usepackage{float}
\usepackage{graphicx}
\usepackage{tcolorbox}
\usepackage{MnSymbol,bbding,pifont}
\usepackage{colortbl}
\usepackage{enumitem}
\usepackage{multirow}
\usepackage{comment}
\usepackage{fontawesome5}
\usepackage{todonotes}
\usepackage{soul}
\usepackage{xcolor}
\usepackage{ragged2e}
\usepackage{array}
\usepackage{tabularx}
\usepackage{rotating}
\usepackage{arydshln}
\usepackage{caption} 
\usepackage{pdfpages}

\usepackage{fontawesome5}
\usepackage{soul}

\newcommand{\highlightcolor}[2]{%
\sethlcolor{#1}%
\hl{#2}%
}

\definecolor{oc-gray-0}{HTML}{F8F9FA}
\definecolor{oc-gray-1}{HTML}{F1F3F5}
\definecolor{oc-gray-2}{HTML}{E9ECEF}
\definecolor{oc-gray-3}{HTML}{DEE2E6}
\definecolor{oc-gray-4}{HTML}{CED4DA}
\definecolor{oc-gray-5}{HTML}{ADB5BD}
\definecolor{oc-gray-6}{HTML}{868E96}
\definecolor{oc-gray-7}{HTML}{495057}
\definecolor{oc-gray-8}{HTML}{343A40}
\definecolor{oc-gray-9}{HTML}{212529}
\definecolor{oc-black}{HTML}{000000}
\definecolor{oc-blue-0}{HTML}{E7F5FF}
\definecolor{oc-blue-1}{HTML}{d0ebff}
\definecolor{oc-blue-7}{HTML}{1C7ED6}
\definecolor{oc-blue-8}{HTML}{1971C2}
\definecolor{oc-blue-9}{HTML}{1864AB}
\definecolor{oc-lime-0}{HTML}{F4FCE3}
\definecolor{oc-lime-8}{HTML}{66A80F}
\definecolor{oc-lime-9}{HTML}{5C940D}
\definecolor{oc-green-0}{HTML}{EBFBEE}
\definecolor{oc-green-3}{HTML}{b2f2bb}
\definecolor{oc-green-8}{HTML}{2F9E44}
\definecolor{oc-green-9}{HTML}{2B8A3E}
\definecolor{oc-orange-0}{HTML}{FFF4E6}
\definecolor{oc-orange-8}{HTML}{E8590C}
\definecolor{oc-orange-9}{HTML}{D9480F}
\definecolor{oc-maroon}{HTML}{A61E4D}

\newtcolorbox{promptbox}[2][Prompt]{
colback=black!4!white, 
arc=5pt, 
boxrule=1.1pt,
fonttitle=\bfseries,
title=#1, 
before upper={\small}, 
fontupper=\selectfont\footnotesize, 
colframe=#2, 
} 

\newcommand{\styledtext}[3]{\textcolor{#1}{\highlightcolor{#2}{#3}}}

\newcommand{\huggingface}{\raisebox{-1.5pt}{\includegraphics[height=1.05em]{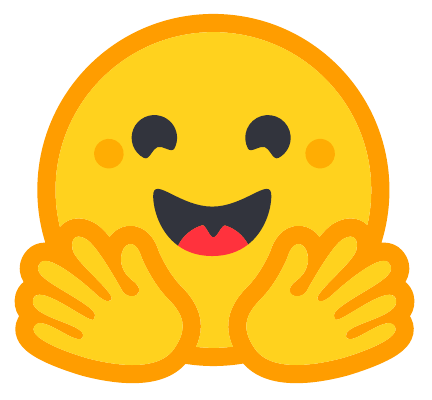}}\xspace}
\newcommand{\github}{\raisebox{-1.5pt}{\includegraphics[height=1.05em]{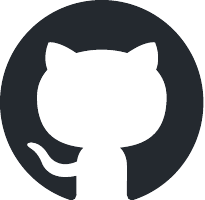}}\xspace}

\iffalse   %% Change between \iffalse and \iftrue to show/hide comments
\newcommand{\yl}[1]{}
\else
\fi

\newcommand{\tulu}{\textsc{T\"ulu}\xspace}
\newcommand{\colorsquare}[1]{\textcolor{#1}{\blacksquare}}
\newcommand{\dataset}{\textsc{SciRIFF}\xspace}
\newcommand{\evaldataset}{\textsc{SciRIFF-Eval}\xspace}

\newcommand{\nsci}{$n_{sci}$\xspace}

\newcommand{\llama}{Llama\xspace}
\newcommand{\qwen}{Qwen 2.5\xspace}
\newcommand{\instruct}{\texttt{-Instruct}\xspace}

\newcommand{\tulumix}{\textsc{T\"ulu V2 Mix}\xspace}
\newcommand{\tuluriffmix}{\textsc{SciRIFF+{T\"ulu}}\xspace}

\newcommand{\tuluname}{\textsc{T\"ulu}\xspace}
\newcommand{\nsciriff}{35,357\xspace}
\newcommand{\ntuluAll}{318,686\xspace}
\newcommand{\nFromScratch}{354,043\xspace}
\newcommand{\nContinued}{70,714\xspace}

\title{\dataset: A Resource to Enhance Language Model Instruction-Following over Scientific Literature\vspace{8pt}}

\newcommand*\samethanks[1][\value{footnote}]{\footnotemark[#1]}

\newcommand{\corethanks}{%
  \begingroup
    \renewcommand\thefootnote{\fnsymbol{footnote}}%
    \footnotemark[2]%
  \endgroup
}

\author{
    David Wadden\thanks{\noindent Equal contribution. Full author contributions \hyperref[sec:contrib]{here}. \\ \hspace{1em}  Correspondence to: \textcolor{BlueViolet}{\{kejian.shi,arman.cohan\}@yale.edu}}\corethanks$^{1}$ \;
    Kejian Shi\samethanks{}\corethanks$^{2}$ \;
    Jacob Morrison\corethanks$^{1,3}$ \;
    Alan Li\corethanks$^{2}$ \;
    Aakanksha Naik\corethanks$^{1}$ \vspace{6pt}\\ 
    \textbf{Shruti Singh}$^{2}$ \;
    \textbf{Nitzan Barzilay}$^{4,1}$ \;
    \textbf{Kyle Lo}$^{1,3}$ \;
    \textbf{Tom Hope}$^{4,1}$ \;
    \textbf{Luca Soldaini}$^{1}$ \vspace{4pt}\\ 
    \textbf{Shannon Zejiang Shen}$^{5}$ \;
    \textbf{Doug Downey}$^{1,6}$ \; \textbf{Hannaneh Hajishirzi}$^{1,3}$ \; \textbf{Arman Cohan}$^{2,1}$ \vspace{6pt}\\
    $^{1}$Allen Institute for AI \quad \quad
    $^{2}$Yale University \\
    $^{3}$University of Washington \quad
    $^{4}$Hebrew University \quad$^{5}$MIT \quad 
    $^{6}$Northwestern University
    \vspace{4pt}\\
    \github \url{https://github.com/allenai/SciRIFF} \\
    \huggingface \url{https://huggingface.co/datasets/allenai/SciRIFF} \\
     \;\; 
}

\begin{document}
\maketitle

\begingroup
\renewcommand\thefootnote{\fnsymbol{footnote}}
\footnotetext[2]{Denotes core contributors.}
\endgroup

\begin{abstract}
We present \dataset (\textbf{Sci}entific \textbf{R}esource for \textbf{I}nstruction-\textbf{F}ollowing and \textbf{F}inetuning), a dataset of 137K instruction-following instances for training and evaluation, covering 54 tasks. These tasks span five core scientific literature understanding capabilities: information extraction, summarization, question answering, claim verification, and classification. \dataset is unique in being entirely expert-written, high-quality instruction-following dataset for extracting and synthesizing information from research literature across diverse scientific fields. It features complex instructions with long input contexts, detailed task descriptions, and structured outputs. To demonstrate its utility, we finetune a series of large language models (LLMs) using a mix of general-domain and \dataset instructions. On nine out-of-distribution held-out tasks (referred to as \evaldataset), LLMs finetuned on \dataset achieve 70.6\% average improvement over baselines trained only on general-domain instructions. \dataset facilitates the development and evaluation of LLMs to help researchers navigate the rapidly growing body of scientific literature.

\end{abstract}

\section{Introduction}
\label{sec:intro}
LLMs have the potential to advance scientific progress by helping researchers navigate and draw insights from the scientific literature. To accomplish these tasks, LLMs must be able to reliably follow a range of \emph{instructions}---e.g. to extract information, summarize content, or answer questions---when given research articles as input. These instructions are often grounded in entire scientific articles, featuring longer inputs than other typical instruction-following resources in the science domain. In addition, the model's responses may need to be \emph{structured} according to a specific format or schema that supports aggregation for literature review \citep{Marshall2019TowardSR}, or is consumable by software components like augmented reading interfaces \citep{Lo2023TheSR,Palani2023RelatedlySL}. 
For example, when analyzing clinical trials, responses should follow a PICO framework (Population, Intervention, Comparison, Outcome), or when examining methodology papers, follow a standardized format capturing study design, sample size, statistical methods, and key findings, or when performing question answering or fact checking, accompany appropriate evidence for attribution and verification. Such outputs can be represented as \texttt{json} to ensure structured, consistent formatting that enhance both human readability and seamless machine processing (e.g., for claim verification and the input claim \texttt{``Coffee consumption reduces diabetes risk''}, the response could be \texttt{\{{
``verdict'': ``support'',
``evidence'': [``Study A shows 23\% risk reduction'', ``Meta-analysis B confirms protective effect''], ``confidence'': ``moderate''}\}).}

While bespoke models are available for specific scientific literature understanding tasks, models that can flexibly follow instructions in domain-specific settings of science are preferable both for their ease of use (offering a unified input / output interface) and for their ability to generalize to novel applications and settings within that domain.

The general instruction-following capabilities of LLMs have advanced rapidly in recent years, largely due to the availability of general-purpose instruction datasets~\citep{zhang2023-instruction-tuning-for}. 
In addition, some instruction-following resources are available for specific scientific and medical tasks, such as describing the properties of a molecule~\citep{fang2023mol,yu2024llasmol} or answering medical exam questions~\citep{Toma2023ClinicalCA,Han2023MedAlpacaA} (see \S\ref{sec:related_work} for a review).
However, few resources are available for supporting instruction-following for flexible scientific literature understanding capabilities across a range of domains.

\begin{figure*}[t!]
  \centering
  \includegraphics[width=0.95\textwidth]{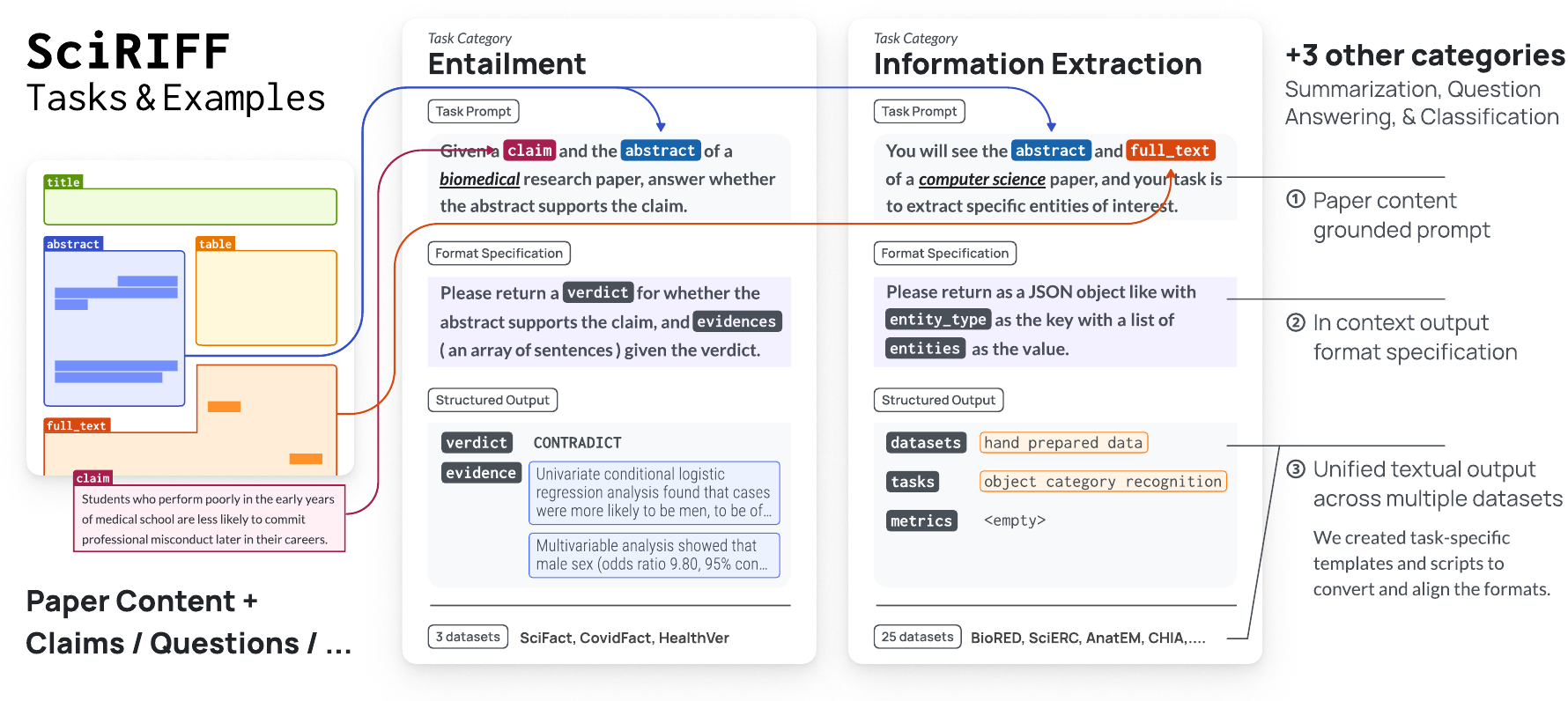}
  \caption{
    Example \dataset tasks. Given an input context from a research paper, the \styledtext{oc-black}{oc-gray-2}{\ \texttt{text prompt}\ } instructs an LLM to perform an operation on the input---e.g. determine whether the \styledtext{oc-gray-0}{oc-blue-8}{\ \texttt{abstract}\ } entails a scientific \styledtext{oc-gray-0}{oc-maroon}{\ \texttt{claim}\ }, extract information over the \styledtext{oc-gray-0}{oc-orange-8}{\ \texttt{full\_text}\ }, answer a question, etc. The model's \styledtext{oc-black}{oc-gray-2}{\ \texttt{output}\ } must conform to a task-specific, user-specified \styledtext{oc-black}{oc-gray-2}{\ \texttt{structure}\ }. \dataset unifies 54 scientific literature understanding tasks under a common input / output format, enabling the development of LLMs that can flexibly generalize to novel scientific use cases.
  }
  \label{fig:task_taxonomy}
\end{figure*}

In this work, we present \dataset (\textbf{Sci}entific \textbf{R}esource for \textbf{I}nstruction-\textbf{F}ollowing and \textbf{F}inetuning), a comprehensive dataset to enable progress on instruction-following over scientific literature. \dataset includes 137K demonstrations for 54 tasks~\ref{sec:appx_full_task_list} spanning five broad scientific literature understanding task categories: information extraction, summarization, question answering, claim verification, and classification. 

\begin{figure*}[t!]
    \centering
    \begin{subfigure}[b]{0.5\textwidth}
        \centering
        \includegraphics[scale=0.26]{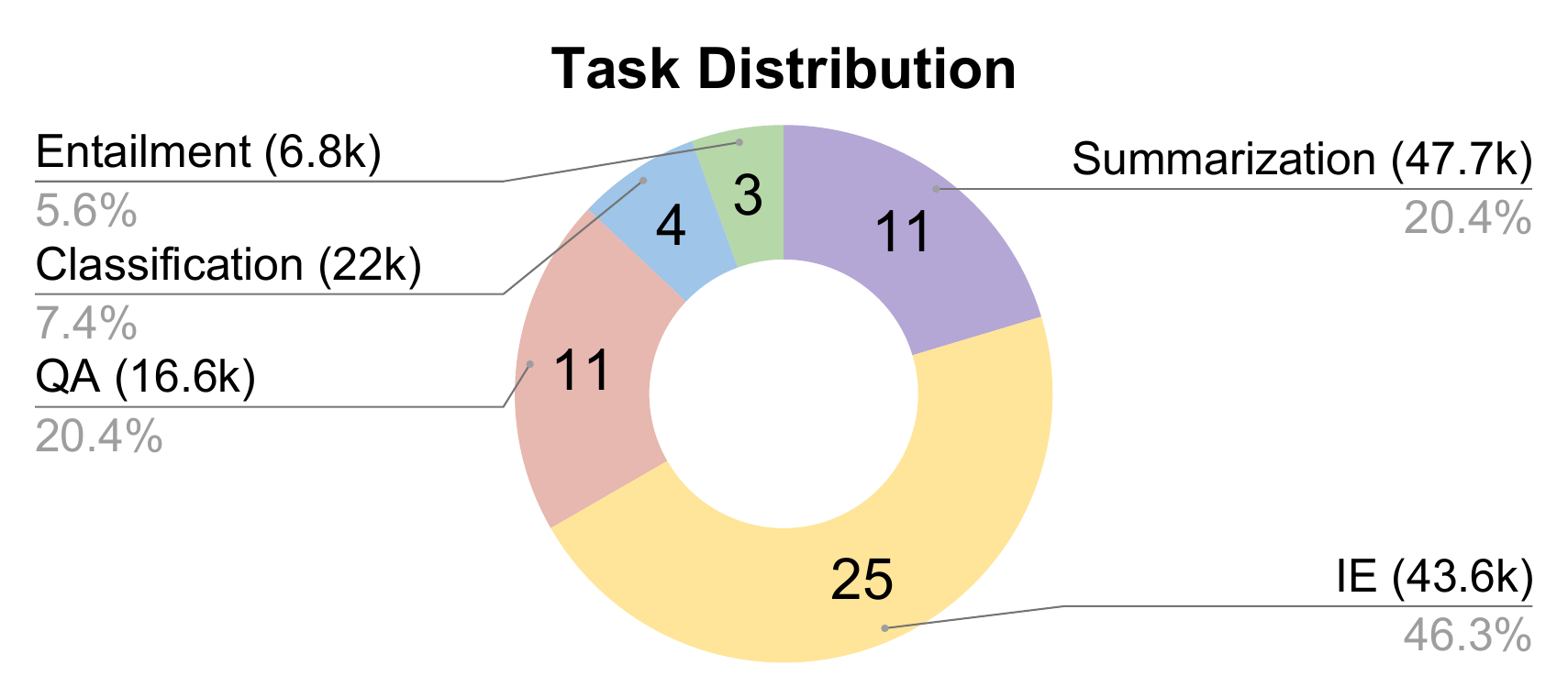}
        \label{subfig:tasks}
    \end{subfigure}%
    ~ 
    \begin{subfigure}[b]{0.5\textwidth}
        \centering
        \includegraphics[scale=0.26]{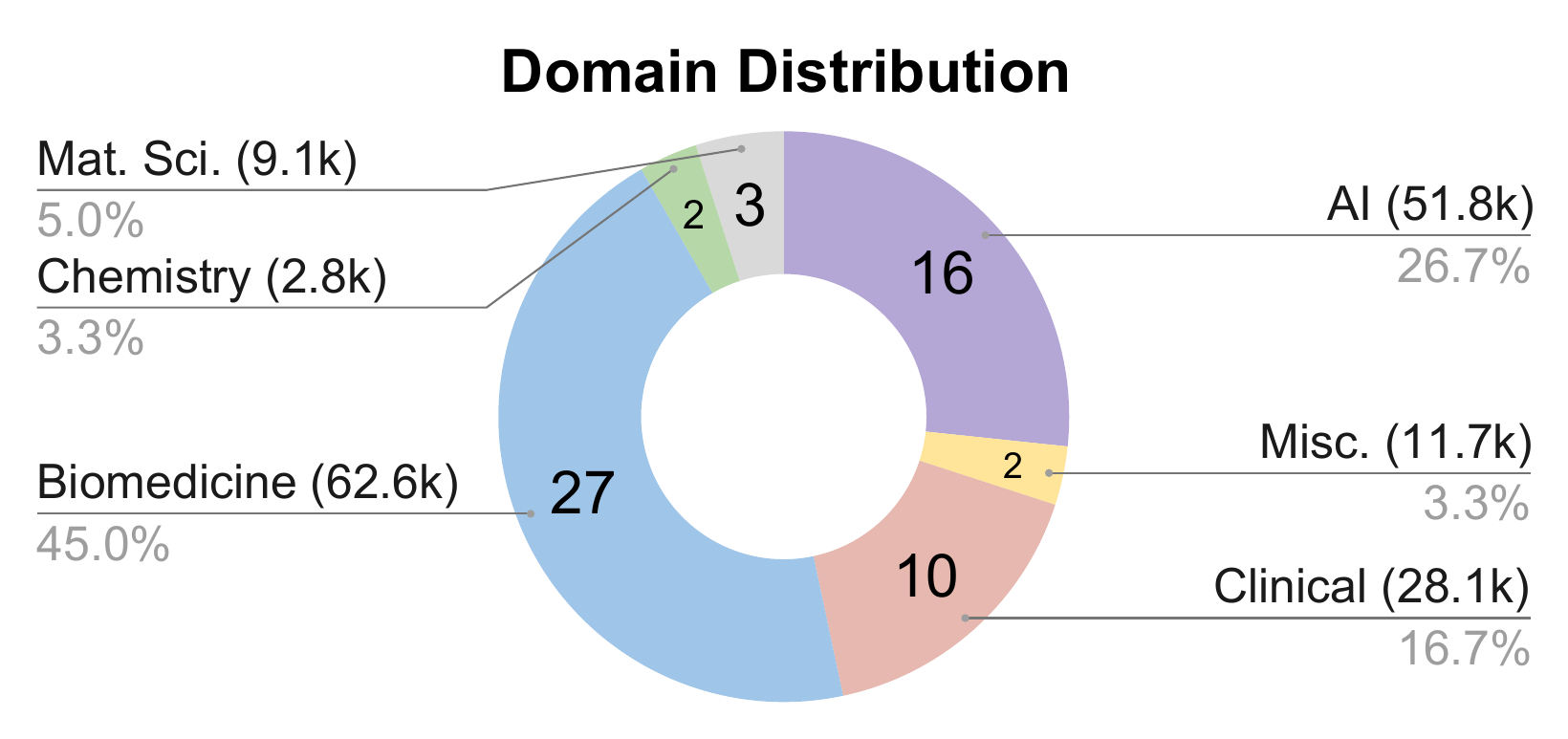}
        \label{subfig:domains}
    \end{subfigure}
    \caption{\dataset: pie charts show dataset counts and brackets indicate instance totals for task categories/domains.}
    \label{fig:dataset_overview}
\end{figure*}

\dataset covers five scientific domains, 
ranging from AI to clinical medicine (Figure \ref{fig:dataset_overview}). 

Unlike synthetic or LLM-distilled instruction-following data (e.g., \citealp{tulu3}), we prioritize human-annotated data to better capture nuanced domain expertise, complex structures, and reasoning required for scientific tasks. Additionally, existing datasets undergo individualized, manually written processes for data conversion to diverse instructions and undergo expert verification, ensuring accuracy and reliability (\S\ref{subsec:dataset_construction}).

Our resource is a unique and specialized instruction-following meta-dataset. As illustrated in Figure \ref{fig:task_taxonomy} and with sample prompt templates provided in Appendix~\ref{sec:appx_prompt_templates}, it is characterized by: (1) grounding every instance in scientific articles or texts, (2) requiring structured and complex responses, such as answers paired with attributions (i.e., tracing the source of the answer), and (3) featuring longer input contexts compared to most existing resources in the science domain (see Figure~\ref{fig:main_token_distribution} and Table~\ref{tab:context_comparison} in Appendix~\ref{sec:appx_sciriff}).

All instruction templates are created by experts (authors of the paper) to ensure quality. Our experiments (\S\ref{sec:results}) show that simple templates—similar to those used in prior work such as FlanV2~\cite{chung2022scaling_flanv2} or generated by an LLM (GPT-4o)—do not capture the complexity of our tasks. As a result, models finetuned on these instructions perform substantially worse than those using our expert-crafted instructions.

We also present a new benchmark dataset \evaldataset (\S\ref{subsec:eval}) for evaluating instruction-following capabilities of LLMs in the science domain. Specifically, we hold out nine datasets from \dataset as an unseen evaluation benchmark which covers a representative range of skills and tasks. To demonstrate the utility of \dataset in improving scientific literature instruction following, we perform supervised finetuning experiments on several LLMs ranging different sizes.\footnote{Other types of post-training such as preference optimization are outside our scope.} When finetuned on a mix of \dataset and general open-source instruction-following data (i.e., Tülu v2 \cite{ivison2023camels_tulu2}), our models show consistent improvements on \evaldataset compared to training on general-domain instructions alone. Our evaluation tasks test true out-of-distribution generalization with formats and templates entirely excluded from training.

In summary, our contributions are as follows:
\
\begin{itemize}[topsep=0pt, align=left, leftmargin=0pt, labelindent=3pt,
listparindent=\parindent, labelwidth=0pt, itemindent=!, itemsep=0pt, parsep=0pt]
\item We introduce \dataset, a high-quality and comprehensive resource for instruction-following in the science domain, containing 137K instances covering a wide range of tasks.
\item We present \evaldataset, a diverse evaluation suite in scientific literature understanding (4.1K selected instances from unseen tasks). 
\item We release a range of LLMs finetuned on \dataset, achieving substantial improvements in scientific literature instruction-following, and conduct experiments showing insights on training strategy and instruction data scaling.
\item We release \dataset dataset, evaluation suite \evaldataset, model checkpoints, and code to enable the community to reproduce our results and contribute to task sourcing for broader coverage.
\end{itemize}

\vspace{-3pt}

\section{\dataset} \label{sec:dataset}

\dataset is a comprehensive instruction-following resource for real-world scientific literature understanding, with 137K instructions for training and in-domain validation. In addition, the test set \evaldataset includes 4.1K instances from held-out tasks.
Our resource spans five task categories and subjects, (Figure~\ref{fig:task_taxonomy}), with particular emphasis on attribution and evidence in scientific tasks. Many tasks require models not only to provide answers but also to support them with evidence from the source paper to ensure verifiable outputs.

% The primary design objective of \dataset is to enhance and evaluate instruction-following capabilities of LLMs in this specialized domain. 
Our focus is on \textit{document-grounded} scientific literature understanding tasks, rather than tasks that evaluate scientific knowledge recall \cite{feng2024sciknoweval}, or general mathematical, problem-solving abilities without reference to scientific literature (e.g., SciInstruct \citep{zhang2024sciglm}, MMLU \citep{hendrycks2021measuring-mmlu}). In addition to a wide coverage, the instructions in \dataset are grounded in long inputs (i.e., paper sections) and support \emph{structured} outputs 
% according to a specific schema 
useful for tasks in literature understanding (such as relation extraction, fact checking with rationale selection, and QA with attribution). \dataset is sourced from existing high-quality scientific datasets and converted into instructions using expert-written and verified instruction templates. Out of 54 tasks, 50 involve templates paired with manually crafted Python scripts that serve to extract ground-truth answers, postprocess (e.g., removing duplicate entity mentions; converting span-level representations to instruction-following formats), or normalize the source datasets.\footnote{The remaining four tasks are naturally formatted for instruction-following and ready for Jinja templating which don't require any special treatment.}

\subsection{Dataset construction} \label{subsec:dataset_construction}

We construct \dataset through a rigorous pipeline that transforms existing scientific literature datasets into high-quality instruction-following instances, which involves template engineering, output schema design, and quality control steps that go well beyond simple dataset reformatting. See \S~\ref{sec:appx_full_task_list} for full task list.

We perform all template writing and annotation with domain experts. Domain experts are the paper authors with extensive experience in NLP. We chose this approach rather than using synthetic data (e.g., \cite{köksal2024longform,li2024selfalignment_humpback}). We believe it is sensible to exhaust available human-annotated resources for this emerging area before turning to potentially noisy synthetic data generation (see Appendix~\ref{sec:appx_prompt_templates} for sample templates, which show the complexity of the tasks.)
Further, in \S\ref{sec:results} we show that using templates from prior work, or using LLMs to generate templates results in significant decline in performance.
In addition, we need high-quality evaluation data, which we construct by holding out nine \dataset tasks as an evaluation benchmark (\S\ref{subsec:eval}). 
% We hope our resource will provide valuable signals for future synthetic data generation efforts.

We adopt \texttt{json} as the output format for structured tasks (34 of 54 tasks).\footnote{Paper authors transform raw dataset annotations into standardized json schemas before templating.} \texttt{json} is a convenient format for tasks requiring attribution, such as information extraction and question answering, where outputs must explicitly pair answers with supporting evidence in a human-and-machine-readable format. Our training set spans multiple scientific domains (Figure~\ref{fig:dataset_overview}). We create instruction mixes of varying context lengths.\footnote{We conduct our experiments using \dataset-4096 (hereafter \dataset) due to computational constraints.} We refer readers to Appendix~\ref{subsec:appx_instructions_statistics} for details and statistics.

\paragraph{Dataset selection criteria}
We focus on scientific literature understanding tasks in which the model is given a portion of scientific text as input, and is instructed to produce a response derived directly from the text.  The task families include summarization, reading comprehension, entailment, classification, and information extraction, which are relevant for real-world use cases (e.g., meta-analysis of literature, clinical decision-making, augmented reading). We provide detailed information and citations of all source datasets in Appendix~\ref{sec:appx_sciriff}. We {\em exclude} datasets that require retrieval from document collections (e.g., open-domain QA), since it is unclear how to build instruction-response pairs from them. We also exclude  datasets that assess general reasoning and mathematical problem-solving skills without grounding on scientific literature, such as ScienceQA \citep{lu2022learnsciQA}, SciBench~\citep{wang2024scibench}, and MATH~\citep{hendrycksmath2021} since such resources already exist. Additionally, we only keep datasets that are publicly available, have a permissive license, and are well-documented and actively maintained. See Appendix \ref{sec:appx_full_task_list} for the complete task list.

% \paragraph{Instruction templates}

\textbf{Quality Verification.} Each template was verified by an additional author for clarity and correctness. We will include guidelines and best practices for prompt-writing in the release and aim to promote community contributions to expand \dataset through our open-sourced data collection process.

\section{Experiment setup}
We conduct supervised finetuning experiments to evaluate the effectiveness of \dataset in improving LLM performance on scientific instruction-following tasks across various model families and sizes. Our experiments explore different data configurations and their impact on scientific instruction-following as measured through \evaldataset described in \S\ref{subsec:eval}. 

% Through extensive experiments, we demonstrate that: (1) base models finetuned with \dataset can achieve comparable or superior performance to their proprietary instruction-tuned counterparts, (2) \dataset provides complementary value even when applied to proprietary instruction-tuned models (WHEN WE FINETUNED ON `--instruct' versions of models), and (3) these improvements hold across different model architectures and sizes, while maintaining general instruction-following capabilities shown by general capability evaluation~ Table~\ref{tab:general_capability}.

\subsection{Evaluation} \label{subsec:eval}

We selected a set of nine tasks from \dataset for evaluation, designed to cover a diverse range of task categories and scientific domains. \dataset tests true out-of-distribution generalization with instructions entirely excluded from training. The inputs, outputs, and evaluation metrics for each task are summarized in Table \ref{tab:testing_tasks}. Additional details of evaluation tasks are included in Appendix~\ref{appx:eval_examples}.

\begin{table*}[!t]
\small
\setlength{\tabcolsep}{3pt}
\centering
\begin{tabular}{lllll}
\toprule
\textbf{Name} & \textbf{Type} & \textbf{Input} & \textbf{Output} & \textbf{Metrics} \\
\midrule
BioASQ List QA & QA & Question, paper excerpts & Answer entities & Exact match F1 \\
BioRED & IE(NER) & Biomedical abstract & 6 entity types & Exact match F1 \\
DiSCoMaT & IE(Table) & LaTex table excerpt & Table entries & BLEU score \\
Evidence Inference (EI) & IE(Rel) & Clinical trial abstract & PICO & String overlap F1 \\
MultiCite (MC) & Classification & Citation context & Citation intents & Exact match F1 \\
MuP & Summarization & ML paper full text & Peer review summary & LLM judge similarity \\
Qasper & QA & NLP paper question & Answer / Attribution & LLM judge similarity / Token F1 \\
SciERC & IE(Rel) & CS abstract & 6 entity types & Exact match F1 \\
SciFact & Entailment & Claim, abstract & Verdict / Evidence & Label F1 / Token F1 \\
\bottomrule
\end{tabular}
    \caption{Evaluation tasks included in \evaldataset. ``/'' separators indicate two separate subtasks. We use \texttt{GPT-4o} as our LLM judge and evaluate similarity on a 1-5 scale; see Appendix \ref{appx:eval_examples} for details. }

    \label{tab:testing_tasks}
\end{table*}

\subsection{Scientific Instruction Finetuning}
\label{subsec:training_settings}

Our goal is to adapt pretrained LLMs to the scientific literature domain. We conduct full finetuning experiments using a range of models and data configurations to assess the effectiveness of \dataset.  In \S\ref{subsec:continued_finetune}, we present an additional analysis examining the potential of using \dataset for continual finetuning of instruction-tuned models, exploring a compute-efficient strategy for adaptation.

\paragraph{Data sources} We finetune using two primary datasets: (1) \textbf{\dataset}, \footnote{In our study, we use 70.5K instances for training.} and (2) \textbf{\tulumix} \citep{ivison2023-camels-in-a}, an open-source high-quality general-domain instruction-following dataset that includes demonstrations from various sources, both human-written (e.g., Flan \citep{wei2021finetuned}) and distilled from proprietary LLMs (e.g., ShareGPT\footnote{\url{https://sharegpt.com/}}, Open Assistant\footnote{\url{https://github.com/LAION-AI/Open-Assistant}}). The original \tulumix contains 326,154 examples, including 7.5K scientific literature understanding demonstrations which overlap (i.e. contaminated) with our evaluation set \evaldataset. We remove those 7.5K examples for clean experiments and to avoid contamination with \evaldataset. For all experiments, we consistently use this filtered version and refer to this as \tulumix to maintain controlled finetuning and unbiased evaluations.

\paragraph{Base models} 
We use following base LLMs as starting points: \llama 3.1-8B \citep{touvron2023llama}, \llama 3.2-3B \citep{dubey2024llama}, and \qwen-1.5B~\citep{yang2024qwen2technicalreport}.\footnote{We do not train larger models due to compute constraint. However, as shown in \S\ref{subsec:main_results} improvements are consistent across sizes/families.} While our primary focus is on improving base models, we also experiment with models that have undergone proprietary instruction tuning and preference optimization (``--instruct'' versions)~\citep{ouyang2022training}. Although direct comparisons with ``--instruct'' models are complicated by unknown training details, we show that \dataset can provide additional value even in these cases. We note, however, that our main results and analyses focus on the controlled experiments with base models where we can fully account for all training conditions.

\paragraph{Finetuning data configurations} For each model, we explore three data configurations: (1) \textbf{\tulumix} only, to establish a \textbf{baseline} for general instruction-following; (2) \textbf{\dataset} only, to assess the impact of scientific instruction data in isolation; and (3) \textbf{\tuluriffmix}, combining the general and scientific instruction data. 

% For the combined configuration, we use all available \tulumix demonstrations along all instances from \dataset.

\section{Results} \label{sec:results}

This section discusses our key results and findings.

\subsection{Main Results} \label{subsec:main_results}

\begin{table*}[!t]
    \centering
    \setlength\tabcolsep{1pt}
    \footnotesize
    \renewcommand{\arraystretch}{1.15}
    \begin{tabular}{@{}llcccccccccr@{}}
        \toprule
        \textbf{Model}              & \textbf{Data}          & \textbf{BioASQ} & \textbf{BioR}  & \textbf{DiscMT} & \textbf{EI}   & \textbf{MC}    & \textbf{MuP}   & \hspace{1.em}\textbf{Qasper}\hspace{1em} & \textbf{SciERC} & \hspace{0.9em}\textbf{SciFact}\hspace{1em} & \textbf{\textsc{Avg.}} \\ \midrule
        % --- New models added for camera-ready ---
        GPT-5                       & -  & 47.3            & 66.6            & 72.0             & 25.7           & \textbf{67.9} & \textbf{94.2}  & 62.1 / \textbf{55.9}                               & 44.0             & 73.1 / \textbf{74.0}                               & \textbf{61.1}                  \\
        Gemini-2.5-Pro                  & -  & 45.1            & 65.1           & 71.2             & 23.7           & 63.9           & 92.8           & \textbf{66.6} / 52.0                               & 42.3            & 77.1 / 72.5                               &  59.8                    \\
        DeepSeek-V3.1                      & -  & 48.3            & \textbf{68.0}  & 75.5            & 25.1          & 51.2           & 90.5           & 66.1 / 49.7                               & \textbf{44.1}            & 82.4 / 61.8                               &  59.2                  \\
        Kimi-K2                     & -  & 49.3            & 66.9           & 75.0            & 25.9         & 58.8           & 92.5           & 61.0 / 47.9                               & 42.7            & \textbf{87.6} / 62.0                      &     60.0                \\
        GPT-4o                      & -  & 48.3            & 63.6  & 71.3            & 25.9          & 62.0  & 88.3           & 54.0 / 55.0                  & 40.3   & 85.5 / 70.4            & 60.4          \\
        GPT-4o-mini                 & -  & \textbf{49.6}   & 53.7           & \textbf{75.6}  & \textbf{27.7} & 54.8           & 88.8  & 61.7 / 46.7                      & 33.1            & 82.7 / 63.6                               & 58.0                    \\ 
        \midrule
        SciLitLLM 7B  & -  & \textbf{51.2} & \textbf{76.6} & \textbf{71.0} & \textbf{23.5} & \textbf{70.7} & 67.5 & \textbf{50.7} / \textbf{53.9} & 49.8 & \textbf{83.4} / \textbf{67.2} & \textbf{60.3}\\ 
        BioMedical-Llama3 8B & -  & 41.1 & 45.7 & 62.9 & 8.4  & 28.6 & 79.8 & 19.0 / 11.1 & \textbf{58.0} & 43.1 / 38.7 & 41.1 \\
        BioMistral 7B           &  -  & 38.3 & 0.7  & 4.7 & 7.7  & 23.7 & 70.3 & 14.1 / 12.5 & 0.0  & 7.1 / 18.6  & 19.1\\
        CodeLlama 7B    & -  & 38.6 & 22.7 & 45.0 & 11.0 & 38.9 & \textbf{80.3} & 46.3 / 31.4 & 14.8 & 55.8 / 35.1 & 38.1 \\
        Llama 2 7B        & - & 34.2 & 0.0    & 4.8  & 7.4  & 37.8 & 72   & 15.7 / 8.5  & 0.3  & 27.7 / 6.2   & 19.5 \\
        \midrule
        \qwen 1.5B\instruct        & -  & 38.9            & 19.7           & 35.5            & 10.5          & 36.9           & \textbf{80.8}  & 38.8 / 39.4                               & 20.8            & 55.0 / 31.5                               & 37.1                    \\
                                     & SciRIFF             & 48.1           & 79.7           & \textbf{80.6}   & 20.9          & \textbf{70.9}  & 67.3           & 42.8 / \textbf{54.3}                      & \textbf{52.0}  & \textbf{80.9} / 68.9                      & \textbf{60.6}          \\
                                     & SciRIFF +Tülu          & \textbf{49.3}   & \textbf{80.1}  & 79.5            & \textbf{21.3} & 70.8           & 61.3           & \textbf{45.8} / 48.6                      & 51.0            & 78.5 / \textbf{70.1}                      & 59.7                    \\
        \noalign{\vskip 0.4ex}\hdashline\noalign{\vskip 0.4ex}
        \qwen 1.5B                  & Tülu         & 35.7            & 23.4           & 31.8            & 7.6           & 6.6            & \textbf{73.0}  & 25.0 / 23.2                               & 12.0            & 52.4 / 29.5                               & 29.1                    \\
                                     & SciRIFF      & 43.6            & \textbf{81.8}  & 45.6            & 18.9          & \textbf{71.2}  & 67.8           & \textbf{47.0} / \textbf{51.4}            & \textbf{52.7}   & 78.8 / 70.5                               & 57.2                    \\
                                     & SciRIFF +Tülu          & \textbf{46.5}   & 79.0           & \textbf{78.3}   & \textbf{19.4} & 70.2           & 63.8           & 40.4 / 49.7                               & 51.7            & \textbf{80.9} / \textbf{70.6}            & \textbf{59.1}          \\ \midrule
        \llama 3.2 3B\instruct     & -  & 42.9            & 35.9           & 61.0            & 11.2          & 47.3           & \textbf{86.0}  & 43.9 / 35.8                               & 20.8            & 59.5 / 40.0                               & 44.0                    \\
                                     & SciRIFF               & 42.7            & \textbf{84.0}  & \textbf{83.4}   & \textbf{25.5} & 71.4           & 64.8           & 50.0 / 57.1                               & \textbf{58.2}   & \textbf{86.8} / \textbf{70.5}            & \textbf{63.1}          \\
                                     & SciRIFF +Tülu          & \textbf{43.0}   & 83.3           & 82.9            & 21.7          & \textbf{72.2}  & 69.0           & \textbf{51.9} / \textbf{58.2}            & 53.3            & 85.6 / 70.3                               & 62.8                    \\ 
        \noalign{\vskip 0.4ex}\hdashline\noalign{\vskip 0.4ex}
        \llama 3.2 3B               & Tülu         & 35.5            & 30.1           & 46.7            & 3.1           & 44.0           & 75.6  & 47.4 / 34.4                      & 20.3            & 55.4 / 36.6                               & 39.0                    \\
                                     & SciRIFF     & 43.6            & 84.2           & 83.2            & \textbf{25.2}          & 71.7           & 64.3           & 46.0 / \textbf{57.2}                      & \textbf{57.2}   & 81.6 / 65.8                               & 61.8                    \\
                                     & SciRIFF+Tülu          & \textbf{46.0}   & \textbf{84.3}  & \textbf{83.3}   & 24.6 & \textbf{72.7}  & 65.5           & \textbf{47.7} / 56.3                               & 57.0            & \textbf{82.7} / \textbf{71.2}            & \textbf{62.8}          \\ 
                                     \midrule
        \llama 3.1 8B\instruct     & -  & 43.7            & 48.8           & 62.2            & 17.8          & 48.8           & \textbf{88.3}  & \textbf{54.0} / 43.0                      & 30.6            & 66.7 / 51.7                               & 50.5                    \\
                                     & SciRIFF             & 45.9            & \textbf{86.0}  & \textbf{83.7}   & 25.0          & \textbf{71.4}  & 70.5           & 53.3 / 54.1                               & \textbf{56.8}   & \textbf{85.8} / \textbf{72.5}            & \textbf{64.1}          \\
                                     & SciRIFF+Tülu          & \textbf{48.8}   & 84.7           & 83.6            & \textbf{26.6} & 71.3           & 66.0           & 50.9 / \textbf{55.2}                      & 54.4            & 85.5 / 70.2                               & 63.4                    \\
        \noalign{\vskip 0.4ex}\hdashline\noalign{\vskip 0.4ex}
        \llama 3.1 8B               & Tülu         & 44.4            & 42.8           & 51.8            & 1.1           & 39.4           & \textbf{80.8}  & 42.8 / 28.6                               & 24.3            & 50.0 / 33.6                               & 40.0                    \\
                                     & SciRIFF     & \textbf{46.2}   & 84.2           & \textbf{83.9}   & 23.5          & 71.0           & 68.5           & \textbf{49.8} / 52.2                      & 56.2            & 83.3 / \textbf{71.9 }                              & 62.8                    \\
                                     & SciRIFF+Tülu          & 41.6            & \textbf{85.2}  & 78.7            & \textbf{28.2} & \textbf{71.6}  & 70.5           & 47.9 / \textbf{61.0}                      & \textbf{58.1}   & \textbf{87.4} / 71.2            & \textbf{63.8}          \\ \bottomrule
    \end{tabular}
    \caption{
             Performance on \evaldataset tasks across model families and training configurations (\S\ref{subsec:training_settings}).  Best performance per model group is \textbf{bolded}. Columns with a ``/'' indicate two evaluation metrics as described in \S \ref{subsec:eval}.
    }
    \label{tab:main_results}
\end{table*}

We report our main experimental results in Table~\ref{tab:main_results}. For fair comparison, all models are finetuned on the same data mixes. 
We show that training on \dataset instances results in the best average performance in each model family. Six frontier models, such as GPT-5, Gemini-2.5-Pro~\citep{gemini2.5} and Kimi-K2~\citep{team2025kimi}, serve as strong baselines.  Additionally, we evaluate selected domain-expert models for comprehensiveness, including SciLitLLM 7B~\cite{li2024scilitllm}, BioMedical-Llama3 8B~\cite{bolton2024biomedlm}, BioMistral 7B~\cite{labrak2024biomistral}, CodeLlama 7B~\cite{rozière2023code}, and a weak baseline Llama 2 7B~\cite{touvron2023llama2}. 

Furthermore, to demonstrate the necessity of expert-written templates for our tasks, we conduct an ablation study comparing our templates against alternatives in \S\ref{subsec:template_ablation}, with details in Appendix~\ref{sec:appendix_template_ablation}.

Our key findings are below:

\paragraph{\dataset enhances scientific literature understanding} 
Table~\ref{tab:main_results} shows that finetuning on SciRIFF consistently enhances the overall performance on \evaldataset. Compared to the corresponding base models finetuned on Tülu, \dataset-trained models achieve, on average, 70.6\% performance gain. Furthermore, without exception, \dataset also adds values when finetuning on ``--instruct'' models (44.6\% on average). Across all model groups, the ``--instruct'' variants trained exclusively on SciRIFF achieve the highest average scores within their respective groups. Finally, while the new frontier models are very strong, with GPT-5 achieving the top baseline score of 61.1, out of the twelve models trained with the inclusion of \dataset instances, eight outperform GPT-5 on \evaldataset, with \qwen 1.5B showing the most significant improvement (from 29.1 to 57.2 in average score with SciRIFF alone). Results indicate that our specialized \dataset can substantially enhance scientific literature understanding and extraction capabilities beyond what general or proprietary instruction data can provide.

% \begin{table}[h]
%     \centering
%     \setlength\tabcolsep{3pt}
%     \footnotesize
%     \renewcommand{\arraystretch}{0.8}
%     \begin{tabular}{@{}l*{6}{c}@{}}
%         \toprule
%         \textbf{Config} & \textbf{L-8B-ours} & \textbf{L-8B-sim} & \textbf{L-8B-syn} & \textbf{Q-1.5B-ours} & \textbf{Q-1.5B-sim} & \textbf{Q-1.5B-syn} \\
%         \midrule
%         Avg. Sci  & 62.8  & 42.2 & --  & 57.2  & 33.1  & --  \\
%         \bottomrule
%     \end{tabular}
%     \caption{Average \evaldataset scores across selected configurations. \texttt{L} stands for Llama-3.1 and Q stands for Qwen-2.5; \texttt{-sim} and \texttt{-syn} represent ``Simple'' and ``Synthetic''; Corresponding columns use their matching evaluations, \evaldataset, \texttt{SciRIFF-Eval-Simple}, and \texttt{SciRIFF-Eval-Synthetic} for fair comparisons (see Appendix~\ref{sec:appendix_template_ablation} for details and the full results at \ksnew{TODO} Table~\ref{}.}
%     \label{tab:template_ablation_main}
% \end{table}
\begin{table}[h]
    \centering
    \setlength\tabcolsep{8pt}
    \footnotesize
    \renewcommand{\arraystretch}{1}
\begin{tabular}{@{}l*{3}{c}@{}}
    \toprule
    \textbf{Config} & \textbf{Ours} & \textbf{Simple} & \textbf{Synthetic} \\
    \midrule
    Llama-3.1-8B      & \textbf{62.8} & 42.2 & 28.0 \\
    Qwen-2.5-1.5B    & \textbf{57.2} & 33.1 & 19.1 \\
    \bottomrule
\end{tabular}
    \caption{Average \evaldataset scores across selected configurations. Columns use their matching evaluations, \evaldataset, SciRIFF-Eval-Simple, and SciRIFF-Eval-Synthetic for fair comparisons. See Appendix~\ref{sec:appendix_template_ablation} and Table~\ref{tab:template_ablation_full} for details.}
    \label{tab:template_ablation_main}
\end{table}

\paragraph{Task-specific impacts and room for improvement} 
\dataset training achieves large gains on the three IE tasks (BioRED, DiSCoMaT, and SciERC). Relative to their Tülu-only counterparts, \dataset-finetuned base models improve IE task performance by, on average, 200.4\%. And SciRIFF training improves performance on QA and Entailment as well. In contrast, performance on the summarization task (MuP) generally decreases after SciRIFF finetuning. This suggests that while SciRIFF is particularly effective for enhancing IE capabilities, it may not provide additional benefits for summarization tasks that are likely well-covered in general instruction-following training. The fact that frontier models our strong finetuned models achieve only an average score of around 60 highlights the difficulty of \evaldataset. Model performance remains relatively low on tasks like EI; This is due to a combination of task difficulty and evaluation challenges, which we discuss in \S\ref{sec:conclusion}.

\paragraph{Balancing scientific and general data}
As shown in Table~\ref{tab:main_results}, combining \dataset and \tulumix training data (\tuluriffmix) yields the best performance on \evaldataset for \textit{base} models. This suggests that incorporating general instruction-following data may provide some broader capability transfer, which base models particularly benefit from, though the impact remains limited (within 2.2\%). On the other hand, training ``--instruct'' models exclusively on SciRIFF data proves to be slightly more effective (within 1\% on average).

\paragraph{Comparing with domain-specialized baselines}
% Models trained on in-domain scientific corpora in continual pretraining, followed by instruction-tuning for science literature tasks, can match GPT-4o on \evaldataset (SciLitLLM). In contrast, models specialized for biomedical and general science tasks (e.g., BioMedical, BioMistral) consistently underperform in literature understanding. Llama 2 7B achieves an average score of only 19.5, with near-zero performance on IE tasks (BioRED and SciERC) partly due to its inability to follow JSON output requirements. We also observe that CodeLlama, likely benefiting from exposure to JSON and code-based reasoning improvements, outperforms Llama 2 and BioMistral. None of the specialized models match the performance of our approach, which uniquely leverages \dataset training to enhance scientific literature understanding.

Models trained on in-domain scientific corpora in continual pretraining, followed by instruction-tuning for science literature tasks, can be very competitive (e.g., SciLitLLM at 60.3 on \evaldataset). In contrast, models specialized for biomedical and general science tasks (e.g., BioMedical, BioMistral) consistently underperform in literature understanding. Llama 2 7B achieves an average score of only 19.5, with near-zero performance on IE tasks (BioRED and SciERC) partly due to its inability to follow JSON output requirements. We also observe that CodeLlama, likely benefiting from exposure to JSON and code-based reasoning improvements, outperforms Llama 2 and BioMistral. None of the specialized models match the performance of our approach, which uniquely leverages \dataset training to enhance scientific literature understanding.

% \subsection{Model Hallucination Analysis} \label{subsec:hallucination}
% \input{tables/hallucination_results}

% We assess whether finetuning on \dataset affects model factuality. We evaluate models on TruthfulQA~\cite{lin2021truthfulqa} and analyze factual consistency in model outputs for the MuP summarization task from \evaldataset. We employ SelfCheckGPT-Prompt~\cite{manakul2023selfcheckgpt} for MuP analysis, using GPT-4o as the judge model (see prompt and metric details in Appendix~\ref{sec:appendix_hallucination}.) Table~\ref{tab:hallucination} shows that models finetuned on \dataset maintain or improve factuality than counterparts trained exclusively on general instruction-following mix.

\paragraph{Grounded Attribution vs. General Reasoning}
While highly capable at strong general reasoning, DeepSeek-V3.1 and Kimi-K2 show lower performance on tasks requiring grounded attribution. Specifically, Table~\ref{tab:main_results} shows their evidence-finding scores on Qasper and SciFact are lower than other frontier models, as is their performance on MultiCite. This suggests a distinction between general problem-solving and the specific skill of finding and attributing evidence from a given text. This finding, also discussed in concurrent work \citep{li2025demystifying}, indicates that strong abstract reasoning does not guarantee proficiency in document-grounded tasks. This reinforces the value of \evaldataset as a specialized benchmark for measuring this crucial, evidence-based capability in scientific literature understanding.

\subsection{Template Ablation}
\label{subsec:template_ablation}

We compare our standard expert-written templates with (1) simple templates that mirror FlanV2~\cite{chung2022scaling_flanv2} and (2) templates generated by GPT-4o. We conduct the analysis on selected (due to compute constraints) \textsc{Base} models with \textsc{SciRIFF} only training data, to exclude confounding factors (see templating details in Appendix~\ref{sec:appendix_template_ablation}.) 
 While prompt ablations are more meaningful for general-purpose language models rather than supervised-finetuned models~\cite{voronov2024mind, instruction-tuning}, we present the experiments to validate our design decisions to rely on expert human-written templates for the emerging and complex domain of instruction-following for scientific literature understanding and synthesis. Table~\ref{tab:template_ablation_main} shows that expert-written templates, which carefully specify task requirements and output structures, outperform the alternatives. We argue, along with detailed descriptions in \S\ref{sec:dataset} and prompt examples at Appendix~\S\ref{sec:appx_prompt_templates}, that expert-written template is preferred. These ablations, while not central to our main contributions and objectives, provide signals on the importance of careful template design for scientific literature understanding tasks.

\subsection{Continual Finetuning Analysis}
\label{subsec:continued_finetune}

In early phase of our study, we explore strategies for efficient adaptation. Specifically, we examined whether starting from an existing instruction-tuned checkpoint (on general domain instructions) could provide compute advantages over training from scratch, without hurting \evaldataset performance. For this controlled experiment, we selected two starting points: (1) \llama 2 base and (2) the same model already finetuned on science-decontaminated \tulumix (referred as \tulu V2). We explored different training approaches:For \llama 2 base, we train on all available \tulumix demonstrations, combined with \textit{1000} instances per SciRIFF task, given the \textit{empirical findings} in \S~\ref{subsec:ablations_sciriff_data}. For the \tulu V2 starting point, we perform continual finetuning using 1000 instances per \dataset task, together with a \textit{matching number} (1000) of instances sampled from \tulumix.

\begin{table}[!t]
    \centering
    \footnotesize
    \setlength{\tabcolsep}{3.5pt} 
\begin{tabular}{@{}llcc@{}}
\toprule
\textbf{Model} & \textbf{Data} & \textbf{7B} & \textbf{70B} \\ \midrule
\llama 2       & \tulu         & 36.7               & 47.5                \\
               & \dataset      & \textbf{48.0}      & \textbf{51.1}       \\
               & \tuluriffmix  & 46.0               & 50.8                \\ \midrule
\multirow{2}{*}{\tuluname V2} & \dataset     & \textbf{47.0}      & 48.8                \\
                              & \tuluriffmix & \textbf{47.0}      & \textbf{50.7}       \\ \bottomrule
\end{tabular}
\caption{Comparison of \evaldataset (\textsc{Sci.}) performance for models finetuned from \llama 2 base and \tulu V2 (science-decontaminated).}
\label{tbl:ablations}
\end{table}

Table \ref{tbl:ablations} reports average \evaldataset performance for our two starting checkpoints using three data configurations. Starting from \tulu V2 performs comparably to \llama 2 base while requiring only 20\% of the compute (Table~\ref{tab:instance_counts}). When trained on \tuluriffmix data, models from both starting points achieve similar performance: \tulu V2 is slightly better on science at 7B and nearly identical at 70B. Given that finetuning \tulu V2 requires only 20\% of the data, this highlights a compute-efficient adaptation for scientific domains, aligning with prior findings \citep{dong2023-how-abilities-in, shi2023-specialist-or-generalist}. While our main experiments (\S\ref{subsec:training_settings}) use newer architectures,\footnote{Due to compute constraints, we do not extend this analysis to all models.} this analysis, along with the results in \S~\ref{subsec:ablations_sciriff_data}, illustrates how practitioners can optimize training for \dataset under fixed model architectures.

\begin{table}[h]
    \centering
    \footnotesize
    \begin{tabular}{lrrr}
        \toprule
        \textbf{Checkpoint} & \textbf{\dataset} & \textbf{\tulu-V2} & \textbf{Total} \\
        \midrule
        \llama 2 base                     & \nsciriff                   & \ntuluAll                   & \nFromScratch  \\
        \tulu V2                 & \nsciriff                   & \nsciriff                   & \nContinued    \\
        \bottomrule
    \end{tabular}
    \caption{\dataset and \tulumix\ instances used for finetuning described in \S\ref{subsec:continued_finetune}, with $n_{sci}=1000$.}
    \label{tab:instance_counts}
\end{table}

% \begin{wrapfigure}[21]{r}{0.5\textwidth}
%     \centering
%     \includegraphics[width=0.5\textwidth]{figures/fig_scaling_v2.pdf}
%     \caption{Performance on \evaldataset as a function of \nsci, the number of science instances per task. Performance gains largely saturate by $n_{sci}=1000$. Experiments are done on 7B models.}
%     \label{fig:7b_scaling}
% \end{wrapfigure}

\vspace{-2pt}

\begin{figure}[!t]
    \centering
    \includegraphics[width=0.4\textwidth]{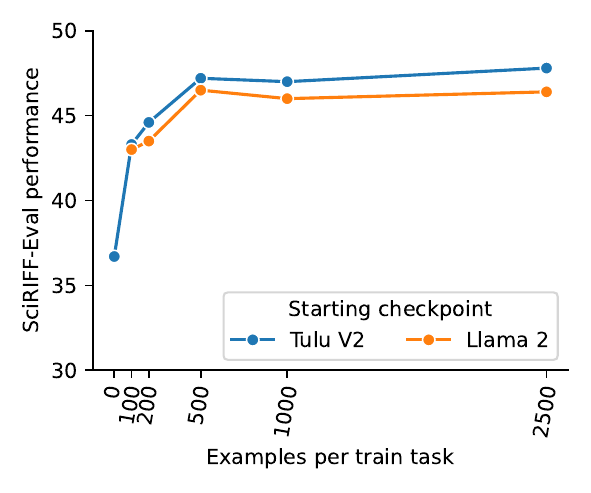}
    \vspace{-1em}
    \caption{
        % Performance on \evaldataset as a function of \nsci, the number of science instances per task. Performance gains largely saturate by $n_{sci}=1000$. Experiments are done on \llama 2 and our \tulu V2 7B models. We extrapolate the finding to settings in \S\ref{subsec:training_settings}.
    Performance on \evaldataset vs. \nsci (instances/task). Gains saturate at $n_{sci}=1000$ (see \S\ref{subsec:training_settings})    
    }
    \vspace{-2pt}
    \label{fig:7b_scaling}
\end{figure}

\subsection{Instruction Data Scale}
\label{subsec:ablations_sciriff_data}

We define \nsci as the number of instances per \dataset task. Figure \ref{fig:7b_scaling} shows that performance on \evaldataset increases sharply as \nsci rises from 100 to 500 and levels off subsequently. We found that 1,000 instances per science task are sufficient for peak performance for \llama 2 models. Therefore, we set $n_{sci}=1000$ across our experiments in the continual finetuning analysis (\S\ref{subsec:continued_finetune}).

\section{Related Work}
\label{sec:related_work}

\paragraph{Strategies for creation of instruction-following resources.} 

Related work has explored a number of methods for curating instruction-following resources such as repurposing existing datasets using human-written templates~\citep{wei2021finetuned, chung2022scaling_flanv2}, crowdsourcing instructions ~\citet{dolly,zhou2023lima, mishra2021-cross-task-generalization}, ShareGPT\footnote{\url{https://sharegpt.com/}} and generating synthetic data \cite{tulu3}. Broadly, synthetic approaches use LLMs to either generate new dataset/task instances alongside instructions \citep{wang2022-self-instruct-aligning, xu2024wizardlm, nayak2024learning, Lou2023MUFFIN}, or to ``back-translate'' existing datasets into instructions \citep{yin2023-dynosaur-a-dynamic, köksal2024longform, li2024selfalignment_humpback}. In this work, we create instructions using human-written templates (\S \ref{subsec:dataset_construction}) for quality assurance. We refer the readers to see template examples in Appendix~\ref{sec:appx_prompt_templates} for evidence. 

\paragraph{Instruction-following resources for scientific literature.} Despite many instruction-following collections, few resources focus on scientific literature, which are crucial for assisting researchers and accelerating discovery \citep{taylor2022-galactica-a-large, xie2023-darwin-series-domain}. Recent work has taken steps in this direction with the development of instruction-following datasets for specific domains such as mathematics \citep{yue2024mammoth,yue2024mammoth2,shao2024-deepseekmath-pushing-the, luo2023wizardmath, tang2024mathscale, toshniwal2024openmath1}, medicine \citep{parmar2022-in-boxbart-get, wu2023-pmc-llama-further, rohanian2023exploring_medtuned}, chemistry \citep{yu2024llasmol, zhang2024-chemllm-a-chemical}, molecular biology \citep{fang2023mol, tran2023-bioinstruct-instruction-tuning}, materials science \citep{song2023-honeybee-progressive-instruction}, and college-level foundational science ~\citep{zhang2024sciglm}. 
% Besides domain limitations, these resources primarily focus on improving LLMs' abilities to solve college-level science problems or reasoning tasks (see also, MMLU~\citep{hendrycks2021measuring-mmlu}, SciEval~\citep{sun2023scieval}, TheoremQA~\citep{chen2023theoremqa}, SciBench~\citep{wang2024scibench}, and GPQA~\citep{rein2023gpqa}).
In contrast, \dataset both covers a broader set of scientific domains and focuses on \textit{document-grounded} scientific literature understanding tasks that can power real-world scientific use cases. 
While recent work such as \citet{li2024scilitllm} explores improving language models' scientific understanding through continuous pretraining and SFT, our work specifically contributes a diverse, high-quality instruction dataset for this domain.
Some instruction-tuning resources have explored structured output formats ~\citep{zhang2023tablellama, wang2023instructuie, jiao2023-instruct-and-extract, gao2023jsontuning}, but not with a focus on science. Finally, most datasets in \dataset have longer instruction contexts than prior works (see Appendix Table~\ref{tab:context_comparison} for a comparison).

\paragraph{Other scientific literature benchmarks.}
Prior works have developed benchmarks to improve and assess scientific literature understanding. Notable efforts in the biomedical domain include BLUE~\citep{peng2019blue-benchmark}, BLURB~\citep{gu2021-blurb}, InBoXBART \citep{parmar2022-in-boxbart-get}, and BigBio~\citep{fries2022bigbio}; \dataset covers a broader set of domains than these resources. Other efforts such as 
~\citep{singh-etal-2023-scirepeval, taylor2022-galactica-a-large, wei2023academicgpt} cover domains beyond biomedicine, but are not targeted for training instruction-following models.
% SciRepEval~\citep{singh-etal-2023-scirepeval}, Galactica~\citep{taylor2022-galactica-a-large}, and AcademicGPT~\citep{wei2023academicgpt} cover domains beyond biomedicine, but are not suitably formatted for training or evaluating instruction-following models.
SciASSESS~\citep{cai2024sciassess} evaluates LLMs' proficiency in scientific literature analysis, focusing on memorization and reasoning abilities. Complementary to our static benchmark, SciArena \citep{zhao2025sciarena} provides a dynamic platform that evaluates models via ongoing expert preference voting.
% \citet{li2024scilitllm} introduce a hybrid strategy that combines continual pretraining and supervised finetuning to specialize LLMs for scientific literature understanding, along with SciLitIns - a \textit{synthetically} generated instruction dataset. 
In contrast, \dataset provides fully \textit{expert-written} instructions, serving both as a benchmark and training resource.

Concurrent with our work, \citet{li2025demystifying} introduce \textsc{SciReas}, a meta-benchmark for scientific problem-solving that includes a subset of \dataset tasks. Their analysis characterizes \dataset as focusing on grounded literature comprehension, distinguishing it from abstract reasoning benchmarks. This distinction is supported by their findings that performance on \dataset has low correlation with reasoning-focused benchmarks like GPQA~\citep{rein2023gpqa}, validating the unique contribution of our resource for measuring essential skills in evidence-based literature understanding.

\section{Conclusion and Future Work} \label{sec:conclusion}

In this work, we introduced \dataset, a resource to facilitate progress on LLM instruction-following over scientific literature. We demonstrated that training on \dataset leads to significant improvement in model performance on held-out scientific tasks (on average 70.6\% over baselines). The large improvements we observe, especially on tasks requiring structured extraction and evidence-finding, underscore the value of targeted data for building practical tools for researchers.

As observed in \S \ref{subsec:main_results}, neither our best finetuned models nor the proprietary frontier models are sufficiently strong on \evaldataset (around 60\%), which demonstrates the difficulty of our tasks. Utilizing LLMs to perform more flexible and finegrained evaluations \citep{kim2023prometheus} represents a promising direction. Future work could focus on reliably generating multiple templates for such complex tasks in a more controlled and principled manner to help models improve their generalization to unseen tasks. Incorporating reliable synthetic data generation techniques and preference data~\citep{lambert2024rewardbench} for scientific literature understanding tasks is also a promising avenue. In conclusion, we are optimistic that the \dataset data and evaluations \evaldataset, as well as the model checkpoints, will serve as valuable resources to build systems for scientific researchers.

\section*{Limitation}
While we demonstrated the effectiveness of \dataset and the value of \evaldataset, we note the following limitations about our work: 
Although we included a wide range of datasets, this still could limit the open-ended tasks that could involve literature understanding. For example, more sophisticated iterative or chat-style interactions mimicking interactions with a research assistant are not captured with \dataset. 
Finally, computational constraints prevented us to experiment with largest open-source models; we suspect that training larger open-source models (such as Llama 3.1 405B) can provide even further improvements over state-of-the-art commercial models. 

\section*{Ethics Statement}
The ethical risks associated with this work are minimal. As we source the data from existing datasets and we work in the science domain, we do not suspect major risks are involved in the curation of our dataset. However, potential biases might still exist in some datasets. For example, one of the source datasets is paper summarization which is sourced from OpenReview.net peer reviews by the original authors. And peer reviews might inherently occasionally include biases or unhelpful languages. As with all LLMs, our trained models are still prone to issues such as hallucinations, so users should exercise caution when interpreting model outputs, particularly in downstream applications in science, and verify any generated content for accuracy and relevance.

\section*{Author contributions}
\label{sec:contrib}
David Wadden and Kejian Shi contributed equally and led the project.
Jacob Morrison, Alan Li, and Aakanksha Naik were among the core contributors and substantially contributed to the experiments and data collection.
Shruti Singh, Nitzan Barzilay, Kyle Lo, Tom Hope, and Luca Soldaini contributed ideas and provided additional support with experiments.
Shannon Shen, Doug Downey, Hanna Hajishirzi, and Arman Cohan provided core mentorship and advising.

\section*{Acknowledgements}
We thank Minyi Chen, Yicheng Gao, Kaiyuan Guan, and Yujie Qiao for their data sourcing contributions during early phase of the project. We are grateful to Google's TRC program for compute support.

\bibliography{biblio, custom}

\appendix

\definecolor{AI}{HTML}{ACD8C9}
\definecolor{Biomed}{HTML}{93B0DA}
\definecolor{Clinic}{HTML}{DDE7F4}
\definecolor{Mat}{HTML}{D7BAE5}
\definecolor{Chem}{HTML}{FDDFD8}
\definecolor{Misc}{HTML}{FDE2AF}
\definecolor{AffilColor}{HTML}{265ed4}

\section{\dataset Provenance}
\label{sec:appx_sciriff}

In this section, we provide additional details for \dataset introduced in the main body of our paper (\S\ref{sec:intro}, \S\ref{sec:dataset}).

\subsection{\dataset Task and Schema}
\label{sec:appx_full_task_list}

% Redefine underscore to allow line breaks
\DeclareRobustCommand{\_}{\textunderscore\hspace{0pt}}

% Define a new column type for left-aligned text in a fixed-width box
\newcolumntype{P}[1]{>{\RaggedRight\arraybackslash}p{#1}}

\begin{table*}[htbp]
\footnotesize
\setlength{\tabcolsep}{12pt}
\centering

\begin{tabular}{P{4cm} l l l}
\toprule
\textbf{SciRIFF name} & \textbf{Source paper} & \textbf{License} & \textbf{Website} \\
\midrule
\texttt{acl\_arc\_intent\_classification} & \href{https://aclanthology.org/L08-1005/}{ACL ARC~\cite{bird-etal-2008-acl}} & - & \href{https://github.com/allenai/scicite/}{[Link]} \\
\texttt{anat\_em\_ner} & \href{https://academic.oup.com/bioinformatics/article/30/6/868/285282}{AnatEM~\cite{pyysalo2014anatomical}} & CC BY & \href{https://nactem.ac.uk/anatomytagger/#AnatEM}{[Link]} \\
\texttt{annotated\_materials\_syntheses\_events} & \href{https://aclanthology.org/W19-4007/}{MatSci Text Corpus~\cite{mysore-etal-2019-materials}} & MIT & \href{https://github.com/olivettigroup/annotated-materials-syntheses}{[Link]} \\
\texttt{bc7\_litcovid\_topic\_classification} & \href{https://pubmed.ncbi.nlm.nih.gov/36043400/}{LitCOVID~\cite{litcovid}} & - & \href{https://biocreative.bioinformatics.udel.edu/tasks/biocreative-vii/track-5/}{[Link]} \\
\texttt{bioasq\_\{factoid,general,list,yesno\}-qa} & \href{https://bmcbioinformatics.biomedcentral.com/articles/10.1186/s12859-015-0564-6}{BioASQ~\cite{bioasq}} & CC BY & \href{http://bioasq.org/}{[Link]} \\
\texttt{biored\_ner} & \href{https://academic.oup.com/bib/article/23/5/bbac282/6645993}{BioRED~\cite{biored}} & - & \href{https://ftp.ncbi.nlm.nih.gov/pub/lu/BioRED/}{[Link]} \\
\texttt{cdr\_ner} & \href{https://www.ncbi.nlm.nih.gov/pmc/articles/PMC4860626/}{BioCreative V CDR~\cite{CDR}} & - & \href{https://biocreative.bioinformatics.udel.edu/tasks/biocreative-v/track-3-cdr/}{[Link]} \\
\texttt{chemdner\_ner} & \href{https://jcheminf.biomedcentral.com/articles/10.1186/1758-2946-7-S1-S2}{CHEMDNER~\cite{Chemdner}} & - & \href{https://biocreative.bioinformatics.udel.edu/resources/biocreative-iv/chemdner-corpus/}{[Link]} \\
\texttt{chemprot\_\{ner,re\}} & \href{https://www.semanticscholar.org/paper/Overview-of-the-BioCreative-VI-chemical-protein-Krallinger-Rabal/eed781f498b563df5a9e8a241c67d63dd1d92ad5}{ChemProt~\cite{chemprot}} & - & \href{https://biocreative.bioinformatics.udel.edu/news/corpora/chemprot-corpus-biocreative-vi/}{[Link]} \\
\texttt{chemsum\_single\_document\_summarization} & \href{https://aclanthology.org/2023.acl-long.587/}{ChemSum~\cite{chemsum}} & - & \href{https://github.com/griff4692/calibrating-summaries}{[Link]} \\
\texttt{chemtables\_te} & \href{https://arxiv.org/abs/2305.14336}{ChemTables~\cite{chemtable}} & GPL 3.0 & \href{https://huggingface.co/datasets/fbaigt/schema-to-json}{[Link]} \\
\texttt{chia\_ner} & \href{https://www.nature.com/articles/s41597-020-00620-0}{Chia~\cite{ChiaHERE}} & CC BY & \href{https://github.com/WengLab-InformaticsResearch/CHIA}{[Link]} \\
\texttt{covid\_deepset\_qa} & \href{https://aclanthology.org/2020.nlpcovid19-acl.18/}{COVID-QA~\cite{covidqa}} & Apache 2.0 & \href{https://github.com/deepset-ai/COVID-QA}{[Link]} \\
\texttt{covidfact\_entailment} & \href{https://aclanthology.org/2021.acl-long.165/}{CovidFact~\cite{covidfact}} & - & \href{https://github.com/asaakyan/covidfact}{[Link]} \\
\texttt{craftchem\_ner} & \href{https://link.springer.com/chapter/10.1007/978-94-024-0881-2\_53}{CRAFT-Chem~\cite{Craftchem}} & - & \href{https://huggingface.co/datasets/ghadeermobasher/CRAFT-Chem}{[Link]} \\
\texttt{data\_reco\_mcq\_\{mc,sc\}} & \href{https://aclanthology.org/2023.acl-long.573/}{DataFinder~\cite{datafinder}} & Apache 2.0 & \href{https://github.com/viswavi/datafinder/tree/main}{[Link]} \\
\texttt{ddi\_ner} & \href{https://www.sciencedirect.com/science/article/pii/S1532046413001123}{DDI~\cite{ddi}} & CC BY & \href{https://github.com/isegura/DDICorpus}{[Link]} \\
\texttt{discomat\_te} & \href{https://aclanthology.org/2023.acl-long.753/}{DISCoMaT~\cite{discomat}} & CC BY-SA & \href{https://github.com/M3RG-IITD/DiSCoMaT}{[Link]} \\
\texttt{drug\_combo\_extraction\_re} & \href{https://aclanthology.org/2022.naacl-main.233/}{Drug Combinations~\cite{drugcombinations}} & - & \href{https://github.com/allenai/drug-combo-extraction}{[Link]} \\
\texttt{evidence\_inference} & \href{https://aclanthology.org/2020.bionlp-1.13/}{Evidence inference~\cite{evidenceinference}} & MIT & \href{https://evidence-inference.ebm-nlp.com/}{[Link]} \\
\texttt{genia\_ner} & \href{https://aclanthology.org/W04-1213/}{JNLPBA~\cite{jnlpba}} & CC BY & \href{https://github.com/spyysalo/jnlpba}{[Link]} \\
\texttt{gnormplus\_ner} & \href{https://www.hindawi.com/journals/bmri/2015/918710/}{GNormPlus~\cite{GNormPlus}} & - & \href{https://www.ncbi.nlm.nih.gov/research/bionlp/Tools/gnormplus/}{[Link]} \\
\texttt{healthver\_entailment} & \href{https://aclanthology.org/2021.findings-emnlp.297/}{HealthVer~\cite{healthver}} & - & \href{https://github.com/sarrouti/healthver}{[Link]} \\
\texttt{linnaeus\_ner} & \href{https://bmcbioinformatics.biomedcentral.com/articles/10.1186/1471-2105-11-85}{LINNAEUS~\cite{LINNAEUS}} & CC BY & \href{https://sourceforge.net/projects/linnaeus/}{[Link]} \\
\texttt{medmentions\_ner} & \href{https://arxiv.org/abs/1902.09476}{MedMentions~\cite{medmentions}} & CC 0 & \href{https://github.com/chanzuckerberg/MedMentions}{[Link]} \\
\texttt{mltables\_te} & \href{https://aclanthology.org/2020.emnlp-main.692/}{AxCell~\cite{axcell}} & Apache 2.0 & \href{https://github.com/paperswithcode/axcell}{[Link]} \\
\texttt{mslr2022\_cochrane\_multidoc\_summarization} & \href{https://www.ncbi.nlm.nih.gov/pmc/articles/PMC8378607/}{Cochrane~\cite{Cochrane}} & Apache 2.0 & \href{https://github.com/allenai/mslr-shared-task}{[Link]} \\
\texttt{mslr2022\_ms2\_multidoc\_summarization} & \href{https://aclanthology.org/2021.emnlp-main.594/}{MS\textsuperscript{2}~\cite{ms2}} & Apache 2.0 & \href{https://github.com/allenai/mslr-shared-task}{[Link]} \\
\texttt{multicite\_intent\_classification} & \href{https://aclanthology.org/2022.naacl-main.137/}{MultiCite~\cite{multicite}} & CC BY-NC & \href{https://github.com/allenai/multicite}{[Link]} \\
\texttt{multixscience\_multidoc\_summarization} & \href{https://aclanthology.org/2020.emnlp-main.648/}{Multi-XScience~\cite{multixscience}} & MIT & \href{https://github.com/yaolu/Multi-XScience}{[Link]} \\
\texttt{mup\_single\_document\_summarization} & \href{https://aclanthology.org/2022.sdp-1.32/}{MUP~\cite{mup}} & Apache 2.0 & \href{https://github.com/allenai/mup}{[Link]} \\
\texttt{ncbi\_ner} & \href{https://pubmed.ncbi.nlm.nih.gov/24393765/}{NCBI Disease~\cite{NCBIDisease}} & CC 0 & \href{https://www.ncbi.nlm.nih.gov/CBBresearch/Dogan/DISEASE/}{[Link]} \\
\texttt{nlmchem\_ner} & \href{https://pubmed.ncbi.nlm.nih.gov/33767203/}{NLM-Chem~\cite{NLMChem}} & CC 0 & \href{https://ftp.ncbi.nlm.nih.gov/pub/lu/BC7-NLM-Chem-track/}{[Link]} \\
\texttt{nlmgene\_ner} & \href{https://pubmed.ncbi.nlm.nih.gov/33839304/}{NLM-Gene~\cite{NLMGene}} & CC 0 & \href{https://ftp.ncbi.nlm.nih.gov/pub/lu/NLMGene/}{[Link]} \\
\texttt{pico\_ner} & \href{https://aclanthology.org/P18-1019/}{EBM-NLP PICO~\cite{pico}} & - & \href{https://github.com/bepnye/EBM-NLP}{[Link]} \\
\texttt{pubmedqa\_qa} & \href{https://aclanthology.org/D19-1259/}{PubMedQA~\cite{pubmedqa}} & MIT & \href{https://github.com/pubmedqa/pubmedqa}{[Link]} \\
\texttt{qasa\_abstractive\_qa} & \href{https://proceedings.mlr.press/v202/lee23n}{QASA~\cite{qasa}} & MIT & \href{https://github.com/lgresearch/QASA}{[Link]} \\
\texttt{qasper\_\{abstractive,extractive\}\_qa} & \href{https://aclanthology.org/2021.naacl-main.365/}{Qasper~\cite{qasper}} & CC BY & \href{https://allenai.org/data/qasper}{[Link]} \\
\texttt{scicite\_classification} & \href{https://aclanthology.org/N19-1361/}{SciCite~\cite{scicite}} & - & \href{https://allenai.org/data/scicite}{[Link]} \\
\texttt{scientific\_lay\_summarisation\_\{elife,plos\}\_single\_doc\_summ} & \href{https://aclanthology.org/2022.emnlp-main.724/}{Lay Summarisation~\cite{scientificlay}} & - & \href{https://github.com/TGoldsack1/Corpora_for_Lay_Summarisation}{[Link]} \\
\texttt{scientific\_papers\_summarization\_\_single\_doc\_\{arxiv,pubmed\}} & \href{https://aclanthology.org/N18-2097/}{Scientific Papers~\cite{scientificpapers}} & - & \href{https://huggingface.co/datasets/armanc/scientific_papers}{[Link]} \\
\texttt{scierc\_\{ner,re\}} & \href{https://aclanthology.org/D18-1360/}{SciERC~\cite{scierc}} & - & \href{http://nlp.cs.washington.edu/sciIE/}{[Link]} \\
\texttt{scifact\_entailment} & \href{https://aclanthology.org/2020.emnlp-main.609/}{SciFact~\cite{scifact}} & CC BY-NC & \href{https://allenai.org/data/scifact}{[Link]} \\
\texttt{scireviewgen\_multidoc\_summarization} & \href{https://aclanthology.org/2023.findings-acl.418/}{SciReviewGen~\cite{scireviewgen}} & CC BY-NC & \href{https://github.com/tetsu9923/SciReviewGen}{[Link]} \\
\texttt{scitldr\_aic} & \href{https://aclanthology.org/2020.findings-emnlp.428/}{SciTLDR~\cite{scitldr}} & Apache 2.0 & \href{https://github.com/allenai/scitldr}{[Link]} \\
\bottomrule
\end{tabular}
\caption{Overview of source datasets repurposed for SciRIFF~(\S\ref{sec:dataset}). SciRIFF is licensed under \texttt{ODC-By} and is derived from existing scientific literature understanding datasets. \text{\{\}} indicates subsets belonging to the same source.}
\label{tab:full_source_list}
\end{table*}

\begin{figure*}[ht]
\centering
\includegraphics[width=\textwidth]{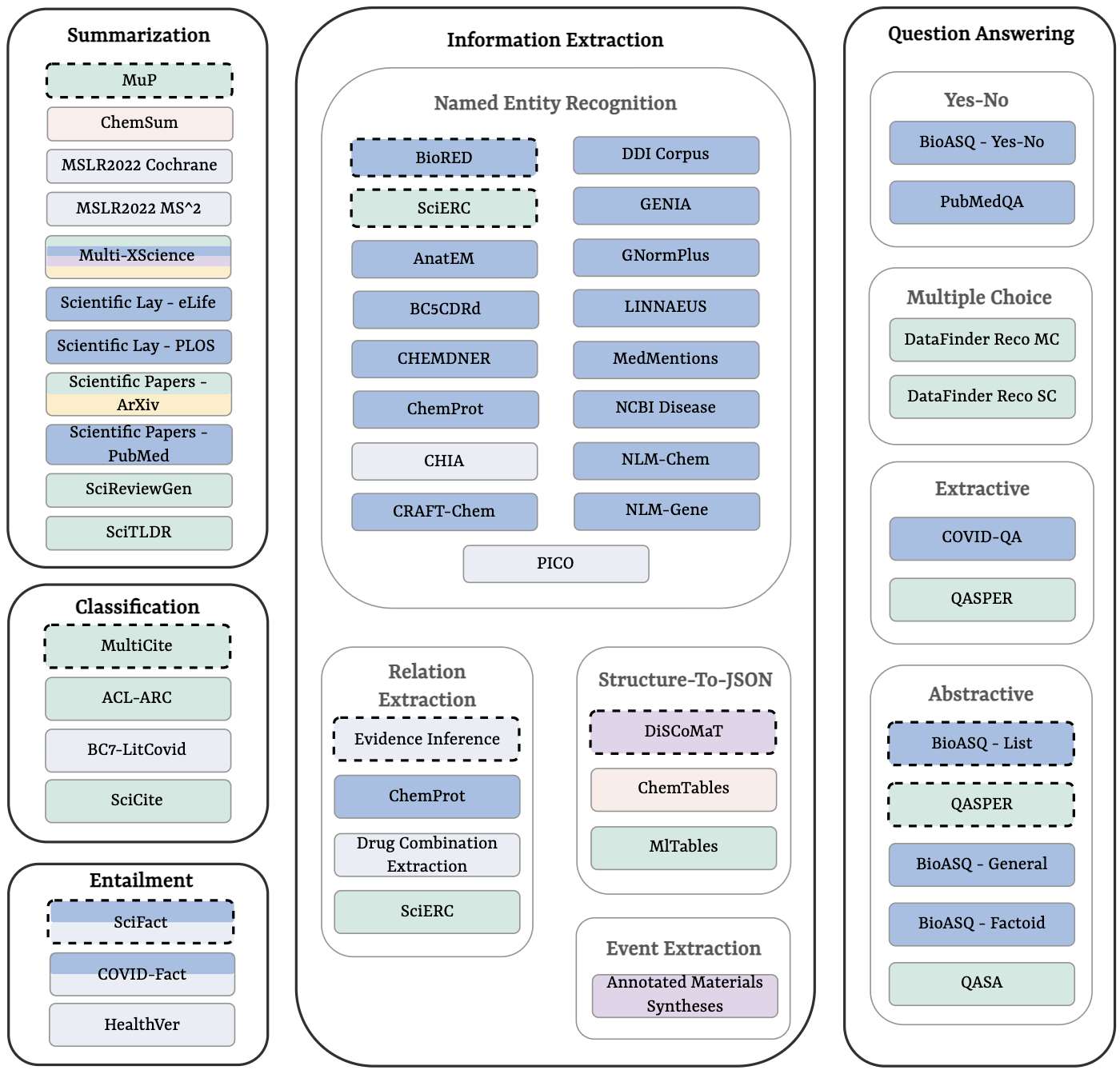}
\caption{Overview of \dataset dataset. Dashed black lines indicate that a task is included in \evaldataset and held out during model training. Scientific domains are colored as follows:
$\colorsquare{Biomed}$Biomedicine;$\colorsquare{AI}$AI;$\colorsquare{Clinic}$Clinical Medicine;$\colorsquare{Chem}$Chemistry;$\colorsquare{Mat}$Materials Science;$\colorsquare{Misc}$Miscellaneous.}
\label{fig:task_list}
\end{figure*}

We provide detailed information on all tasks--including citations, URLs to source websites, and licensing information where available--in Table~\ref{tab:full_source_list}.
\dataset task taxonomy is visualized in Figure \ref{fig:task_list}.  Where convenient, we use datasets as preprocessed by the BigBio resource (\url{https://huggingface.co/bigbio}); details will also be provided in the dataset card upon release.

\subsection{Task Length Distribution} \label{appx:task_length}

\begin{figure*}[t!]
  \centering
  \includegraphics[width=1\textwidth]{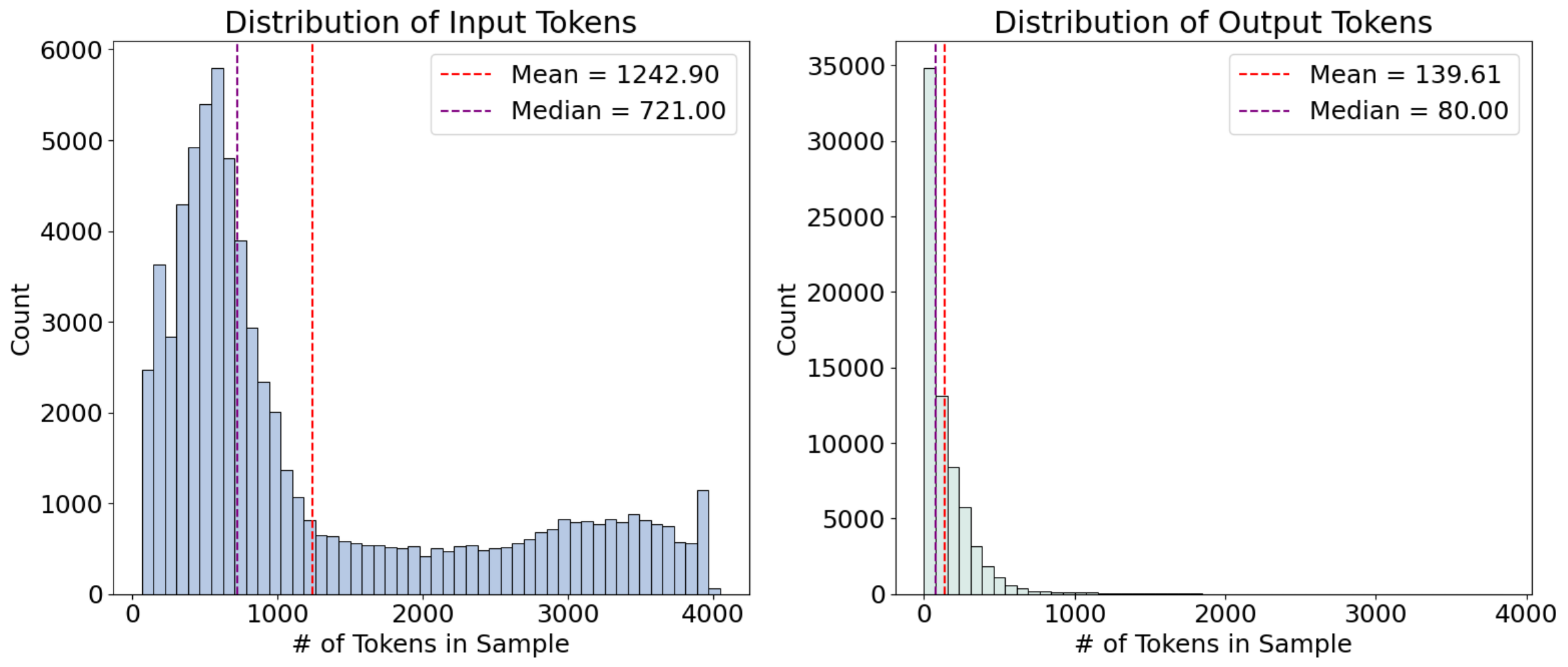}
  \caption{Distribution of input (left) and output (right) token lengths over \dataset training instances.}
  \label{fig:main_token_distribution}
\end{figure*}
\begin{table*}[!t]
    \centering
    \footnotesize

    \begin{tabular}{lccc}
        \toprule
        Name                                                      & \# Instances & Domain                             & Avg. Length             \\
        \midrule[1pt]
        \rowcolor[RGB]{222, 230, 242}
        \multicolumn{4}{l}{\textit{General Domain}}                                                                                             \\
        Flan V2~\citep{chung2022scaling_flanv2}                   & 15M          & General                            & 355.6 / 31.2            \\
        SuperNI~\citep{wang2022supernaturalinstructions}          & 97K          & General                            & 291.1 / 38.7            \\
        \tulumix~\citep{ivison2023-camels-in-a}                   & 326K         & General                            & 353.3 / 696.9           \\

        \midrule[1pt]
        \rowcolor[RGB]{203, 230, 222}
        \multicolumn{4}{l}{\textit{Scientific Domain}}                                                                                          \\
        BoX~\citep{parmar2022-in-boxbart-get}                     & 141K         & Biomed                             & X$^{*}$                   \\
        SciInstruct~\citep{zhang2024sciglm}                       & 254K         & Math, PH, Chem, FP                 & 88.4 / 265.6\           \\
        Mol-Instructions~\citep{fang2023mol}                      & 2.04M        & Biomolecular                       & 126.3 / 112.9           \\
        MathInstruct~\citep{yue2024mammoth}                       & 262K         & Math                               & 82.5 / 174.0            \\
        MedInstruct-52K~\citep{zhang2023alpacareinstructiontuned} & 52K          & Medical                            & 148.2 / 96.9            \\
        LlaSMol~\citep{yu2024llasmol}                             & 3.29M        & Chem                               & 81.9 / 53.0             \\
        \midrule
        \dataset (Our work)                                       & 137K         & AI, Biomed, Clinical, Chem, MatSci & \textbf{1242.9 / 139.6} \\
        \bottomrule
    \end{tabular}
    \caption{Comparison with selected instruction-following datasets. We use the following abbreviations: PH -- Physics; FP -- Formal Proof; MatSci -- Materials Science. We report average token counts for \texttt{input/output} using \llama 2 tokenizer using up to 200k subsamples from each dataset. $^{*}$BoX dataset is not readily available.}
    \label{tab:context_comparison}
\end{table*}

% \footnote{Only templates and description provided. Data not available.}

Figure \ref{fig:main_token_distribution} shows the distribution of input and output lengths for demonstrations in \dataset.

Table \ref{tab:context_comparison} compares \dataset with selected instruction-following datasets, including canonical collections commonly used for general fine-tuning and selected datasets specialized in scientific domains. Our dataset features longer input contexts than existing resources.

\subsection{Instruction Mix Statistics}
\label{subsec:appx_instructions_statistics}

We further describe our data mixture following the main discussion in \S\ref{sec:dataset}. Figure~\ref{fig:dataset_overview} presents an overview of the \dataset training set distribution over task categories and domains. The domain distribution reflects the current landscape of available high-quality scientific  datasets (e.g., \citealp{reid2022m2d2}), with a notable representation from the biomedicine and AI domain. This aligns with our dataset selection criteria, which prioritize well-documented resources with permissive licenses. 

Given the significant presence of information extraction tasks, a large percentage of datasets in \dataset (34 datasets; 63\%) require structured outputs.

We construct three instruction mixes from this dataset collection, with maximum context lengths (input + output tokens) of 4,096, 8,192 and 16,382 per instance (longer instances are truncated where possible and discarded otherwise; see Appendix \ref{subsec:truncation}). Due to model and hardware limitations, we conduct experiments in this work using the \dataset-4096 mixture, and make the longer mixtures available to enable future research. In what follows, we refer to \dataset-4096 simply as \dataset.

\subsection{Truncation Strategy} \label{subsec:truncation}

In \S \ref{subsec:appx_instructions_statistics}, we mention that when an instance exceeds the maximum context length for a given version of \dataset, we truncate where possible and discard otherwise. In particular, we truncate for tasks (like question answering) where the task output can be localized to particular passages in the input document by randomly removing irrelevant passages until the document fits in the desired context. For tasks like summarization, where the task output cannot easily be localized, we simply discard examples that are longer than the context window.

\section{Template Ablation}
\label{sec:appendix_template_ablation}

We created two variants of templates for comparison: (1) simple templates adapted from previous work FlanV2, a collection of datasets, templates, and methods for general-purpose instruction tuning~\citep{chung2022scaling_flanv2}, and (2) LLM-generated templates with GPT-4o. 

\subsection{Evaluation under Alternative Template}
For fair evaluation, we develop corresponding variants of our evaluation templates (for SciRIFF-Eval tasks; \S\ref{subsec:eval}) to ensure that models trained on alternative templates are evaluated on prompts of matching distribution.

\subsection{Simple Template} \label{subsec:appx_simple_template}

We adapted the style of FlanV2's basic instruction format \textbf{while maintaining essential task requirements}. For example, we transformed complex templates into basic input-output patterns (e.g., \texttt{Summarize:{text}\textbackslash n \textbackslash n Summary: \textbackslash n}) while preserving necessary variable substitutions using "{{variable}}" syntax in Jinja. To ensure valid comparison and prevent complete task failure, we maintained minimal but crucial specifications such as output format requirements (e.g., JSON structure) and output constraints where necessary.

\begin{figure*}[t!]
\begin{tcolorbox}[colback=black!3!white, colframe=black!70!white, title=SciERC-simple, fontupper=\footnotesize, fonttitle=\footnotesize]

Extract all unique entities from the paper abstract.\newline

Output a JSON object where keys are entity types and values are lists of extracted entities.\newline

Abstract:\newline

\{\{paper\}\}\newline

|||\newline

\{\{ ner\_dict | tojson \}\}

\end{tcolorbox}
\caption{simple template for \texttt{SciERC} task.}
\label{fig:prompt_simple_scierc}
\end{figure*}

\begin{figure*}[t!]
\begin{tcolorbox}[colback=black!3!white, colframe=black!70!white, title=QASPER-simple, fontupper=\footnotesize, fonttitle=\footnotesize]

Read the following paper excerpts and answer the question. Output a JSON object with "answer" and "evidence" fields. \newline

Paper: \{\{paper\}\}\newline

Question: \{\{question\}\}\newline

|||\newline

\{\{output\}\}\newline

\end{tcolorbox}
\caption{simple template for \texttt{QASPER} task.}
\label{fig:prompt_simple_qasper}
\end{figure*}

\begin{figure*}[t!]
\begin{tcolorbox}[colback=black!3!white, colframe=black!70!white, title=Prompt Generation, fontupper=\footnotesize, fonttitle=\footnotesize]

Today, you will write instruction templates (in Jinja) to format an instruction-following task that a researcher might reasonably ask about scientific literature.\newline

You will be writing templates in Jinja formats. Input field and output field are separated by "|||". Since our Jinja template will likely be a multiline string, please use a block scalar "|" to indicate a multiline string in Jinja. For example:\newline
""" \newline
jinja: |\newline
    <input part: most of your instructions will be in this part>\newline
    |||\newline
    <output part>\newline
"""\newline

Here is the task that you are about to create template for:\newline

\{\{TASK\_DESCRIPTION\}\}\newline

\{\{RELEVANT\_CONTEXT\}\}\newline

<---start\_of\_author\_notes---NOT IN ACTUAL PROMPT>\newline
Author notes: We optionally provide ``relevant context'' in this general format. In actual generation, we drop fields that do not apply.\newline

- task\_family: The category to which this task belongs. Options include summarization, ie, qa, entailment, and classification.\newline
- domain: Scientific field(s) that the task covers like "artificial intelligence"\newline
- input\_context: Whether the input is full paper text, a table, etc.\newline
- source\_type: Indicates whether the input comes from a single paper or multiple sources.\newline
- output\_context: Clear text descriptions for output requirements like "Yes or No",  json, jsonlines.\newline
<---end\_of\_author\_notes---NOT IN ACTUAL PROMPT>\newline\newline

You should clearly and concisely specify task requirements and any special output structures (if applicable). For tasks that require JSON (or JSON array) outputs, explicitly mention the output requirement in your template.\newline

Here is the list of anchor variables for this task, which are prepared for you:\newline
\{\{VARIABLES\_POSTPROCESSED\_BY\_EXPERT\_ANNOTATORS\}\}\newline
\\
Important: the content enclosed by "\{\{" and "\}\}": should NOT change. You should re-use the verbatim texts for anchor variables.\newline\newline
Here is a template example belonging to the same task category. You should only study the overall structure and the style, but do not copy the content:\newline

\{\{DEMONSTRATION\_FROM\_STANDARD\_SCIRIFF\_INSTRUCTION\}\}\newline

Make sure your generated template prompt is clear and not verbose.\newline

\end{tcolorbox}
\caption{Template generation prompt for GPT-4o for synthetic templates \S\ref{appx_synthetic_templates}. We adapt the prompt for individual tasks. We note that GPT-4o often generate vague and under-specified instructions for our use case.}
\label{fig:prompt_gpt4o}
\end{figure*}

\begin{table*}[!t]
    \centering
    \setlength\tabcolsep{1pt}
    \footnotesize
    \renewcommand{\arraystretch}{1.15}
    \begin{tabular}{@{}llccccccccccc@{}}
        \toprule
        \textbf{Model}         & \textbf{Data}  & \textbf{BioASQ} & \textbf{BioR} & \textbf{DiscMT} & \textbf{EI} & \textbf{MC} & \textbf{MuP} & \hspace{1.em}\textbf{Qasper}\hspace{1.em} & \textbf{SciERC} & \hspace{0.9em}\textbf{SciFact}\hspace{1.em} & \textbf{Sci.} & \textbf{Sci.Selected} \\ \midrule
        \llama-3.1-8B & Ours       & 46.2   & 84.2   & 83.9   & 23.5  & 71.0  & 68.5  & 49.8 / 52.2  & 56.2  & 83.3 / 71.9  & 62.8  & 65.0 \\
                      & Simple     & 57.3   & 64.4   & 19.6   & 4.1   & 9.4   & 42.3  & 49.8 / 65.0    & 33.7  & 65.8 / 52.6  & 42.2  & 36.6 \\
                      & Synthetic  & 41.0     & 58.1   & 38.7   & 0.3   & 9.1   & 57.5  & 0.0            & 0.0     & 63.4 / 39.7  & 28.0    & 36.7 \\ \midrule
        Qwen2.5-1.5B  & Ours       & 43.6   & 81.8   & 45.6   & 18.9  & 71.2  & 67.8  & 47.0 / 51.4  & 52.7  & 78.8 / 70.5  & 57.2  & 57.7 \\
                      & Simple     & 36.2   & 43.1   & 38.5   & 0.8   & 2.9   & 42.5  & 32.5 / 48.3  & 28.8  & 47.9 / 42.2  & 33.0    & 29.9 \\
                      & Synthetic  & 33.9   & 39.7   & 40.9   & 0.3   & 7.7   & 68.5  & 0.0            & 0.0     & 7.2 / 11.8   & 19.1  & 28.6 \\ \bottomrule
    \end{tabular}
    \caption{Performance on evaluation tasks (\evaldataset, \texttt{SciRIFF-Eval-Simple}, and \texttt{SciRIFF-Eval-Synthetic} respectively, across ablations for \S\ref{subsec:template_ablation} and Appendix~\ref{sec:appendix_template_ablation}. This table accompanies Table~\ref{tab:template_ablation_main}. Sci.Selected represents the average score dropping QASPER and SciERC tasks (representative of complex output in SciRIFF), where synthetic templates failed to specify the required complex output formats. We show that our templates show stronger performances under either comparison scheme.}
    \label{tab:template_ablation_full}
\end{table*}

The simple templates still require careful adaptation since previous work typically handles simpler scenarios - their templates rarely need to ground instructions in scientific papers or specify structured output formats. We made deliberate choices to preserve these critical requirements while simplifying the instruction language and reducing template complexity. Figures \ref{fig:prompt_simple_scierc} and \ref{fig:prompt_simple_qasper} show examples of simple templates. 

For evaluation, we created \texttt{SciRIFF-Eval-Simple}, a variant of our evaluation suite using simple prompts. This ensures that models trained on simple templates are not unfairly evaluated on complex instructions, while still testing the core capabilities required for scientific literature understanding tasks.

\subsection{Synthetic Template} \label{appx_synthetic_templates}

We also explored using GPT-4o to generate instruction templates\footnote{Initial attempts at naive prompting failed to produce usable templates.}. 

For each task category, we provided GPT-4o with a canonical example template and detailed specifications including task requirements, input-output structures, and available variables (\{\{ anchors\}\}) from our prior post-processing steps (See \S\ref{sec:dataset}). Generating templates for diverse scientific literature understanding tasks proved challenging. The complexity of our tasks--ranging from evidence-based question answering to structured information extraction-- makes it difficult to create a universal prompting strategy.

We provide our prompt template for synthetic template generation in Figure~\ref{fig:prompt_gpt4o}. 

For evaluation, similar to the approach in \S\ref{subsec:appx_simple_template}, we created \texttt{SciRIFF-Eval-Synthetic}, following the same principle of matching training and evaluation distribution. 

\subsection{Results and Discussion} \label{appx_template_ablation_results}

Table~\ref{tab:template_ablation_full} shows that expert-written templates, which carefully specify task requirements and output structures, outperform the alternatives.
Additionally, we observe that the (in-distribution) evaluation for \texttt{Synthetic} variants show zero performance on QASPER and SciERC tasks (See Figure~\ref{fig:prompt_qasper} and Figure~\ref{fig:prompt_scierc}--our expert-crafted template--for reference). Upon inspection, we found that GPT-4o\footnote{Note that the effort to prompt GPT4o to generate template for diverse and different scientific literature understanding tasks is non-trivial in itself; See Figure~\ref{fig:prompt_gpt4o}.} failed to specify the required output format correctly, thus the evaluation fails. Nevertheless, when we drop the two tasks, we still see that expert-written templates perform much stronger than the alternatives.

\section{Training Details}

\label{app:hyperparams}

For instruction-tuning, our training hyperparameters were as follows:
\begin{itemize}[noitemsep]
    \item Precision: BFloat16
    \item Epochs: 5
    \item Weight decay: 0
    \item Warmup ratio: 0.03
    \item Learning rate: 2e-5 (1e-5 for 70B)
    \item Max. seq. length: 4,096
    \item Effective batch size: 128
\end{itemize}

For context, each training run of 7B-sized models requires approximately 40 GPU hours on H100 GPUs, making comprehensive ablation studies (on e.g. task mixing ratios) prohibitively expensive for most research labs. We have prioritized our computational resources for experiments that directly address core research questions while maintaining reproducibility for typical computing budgets.
\section{Evaluation Details}  \label{appx:eval_examples}

% The following pages show full input / output examples for all \evaldataset tasks, along with details on metric calculations. This information will be available on our project GitHub page. For tasks using an LLM judge, we found in preliminary experiments that the results of GPT-3.5 were similar to other proprietary LLMs like GPT-4 and Claude-2; we used GPT-3.5 in the interest of cost and efficiency.

The following pages show full input / output examples for all \evaldataset tasks, along with details on metric calculations. This information will be available on our project GitHub page. We use gpt-4o-2024-08-06 model for tasks using an LLM judge as evaluation.

\clearpage 

\includepdf[pages=-,pagecommand={},width=1.2\textwidth]{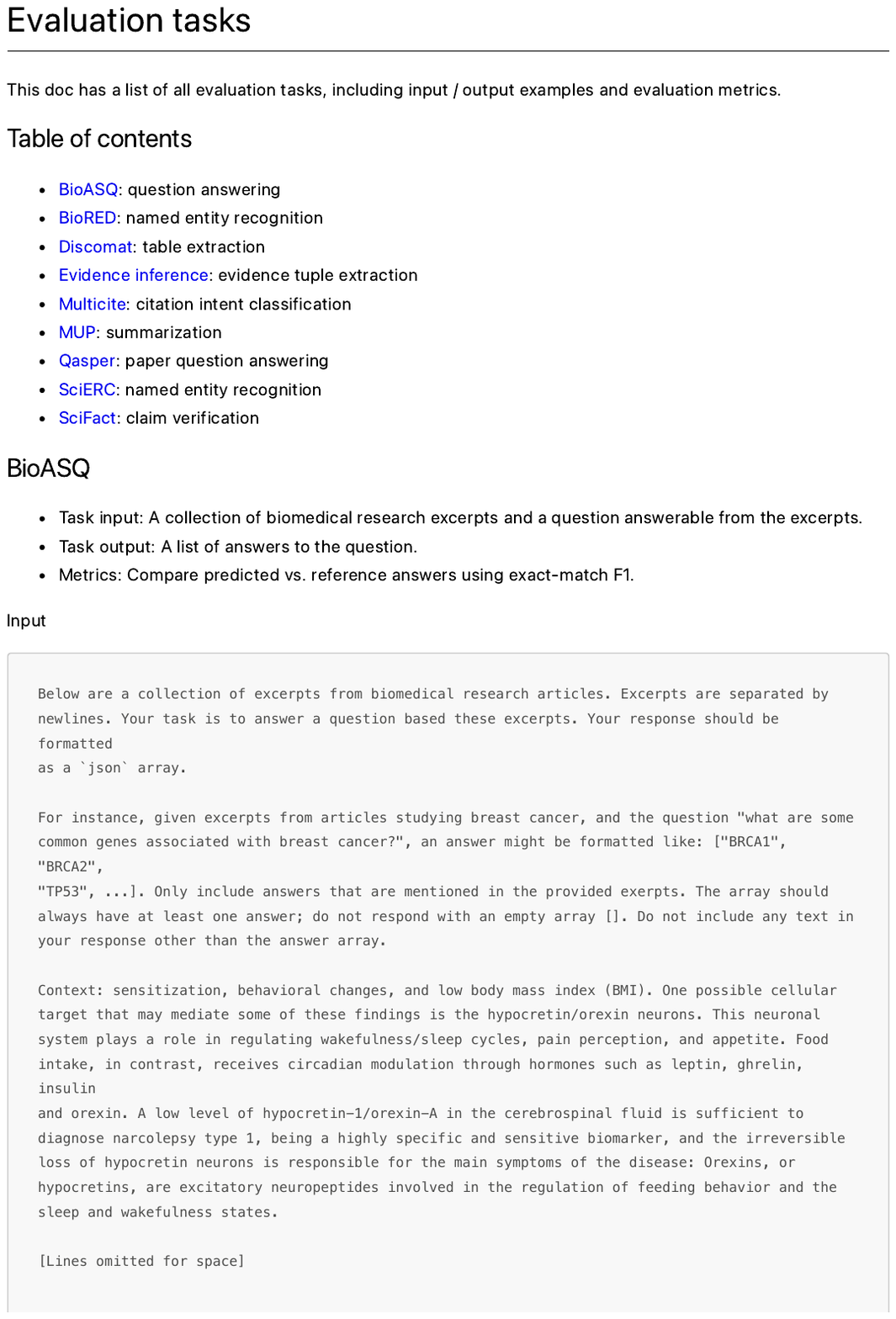}
\section{Instruction Template Creation}
\label{appx:prompt_guidelines}

Instruction templates are written in ~\citep{jinja}, Guidelines and best practices'' for prompt-writing will be  available at our GitHub repository. Each prompt was double-checked by an additional paper author for clarity and correctness.

\section{Sample template} \label{sec:appx_prompt_templates}
In this section, we provide examples of our expert-written templates that demonstrate the complexity and precision required for scientific literature understanding tasks, described in \S\ref{sec:intro} and \S\ref{subsec:dataset_construction}. These templates are carefully designed to elicit structured outputs while requiring sophisticated capabilities such as information extraction with attribution, multi-step reasoning, and adherence to specific output schemas. The templates shown --QASPER (QA, Figure~\ref{fig:prompt_qasper}), SciERC (IE, Figure~\ref{fig:prompt_scierc}), HealthVer (Fact-checking, Figure~\ref{fig:prompt_healthver}), DiSCoMaT (IE over tabular data, Figure~\ref{fig:prompt_discomat}), and DataFinder Reco MC (Multiple Choice QA, Figure~\ref{fig:prompt_datareco}) -- demonstrates how our instruction format guides models to perform challenging tasks like answering questions with evidence attribution, extracting nested entity relationships, and verifying scientific claims with supporting rationales.\footnote{Our preliminary experiments showed that even strong proprietary models like GPT-4o struggled to reliably generate such structured outputs without explicit templates. This observation motivated our decision to use expert-written templates.} 

\begin{figure*}[t!]
\begin{tcolorbox}[colback=black!3!white, colframe=black!70!white, title=QASPER, fontupper=\footnotesize, fonttitle=\footnotesize]

  You will be shown sections from a scientific research paper, together with a question about the paper. This is an extractive question-answering task, where you must find and extract relevant text spans directly from the paper to answer the question. Your response should strictly be a json object with two fields:\newline

  - "answer": An array of strings extracted directly from the paper which, collectively, answer the question.
  
  - "evidence": An array of strings. Each should be an excerpt from the paper, in which one or more of the extracted answers can be found.\newline

  For example, for the question "What baselines did the authors compare against?", a sample response might be:
  
  \{
  
    answer'': [BERT'',RoBERT''],
    
    evidence'': [In our experiments, we compare the performance of our model against BERT and RoBERTa.'']
    
  \}

  Do not include any text in your response other than the json.\newline
  If the question is unanswerable given the provided excerpts, respond with the single word "null".\newline
  Paper: \{\{paper\}\}\newline\newline
  Question: \{\{question\}\}\newline
    
  |||\newline

  \{\% if unanswerable \%\}
  null\newline
  \{\% else \%\}\newline
  \{\{ \{"answer": answer, "evidence": evidence\} | tojson \}\}\newline
  \{\% endif \%\}

\end{tcolorbox}
\caption{Canonical template for \texttt{QASPER} task in Figure~\ref{fig:task_list}. See \S\ref{sec:appx_prompt_templates} for description.}
\label{fig:prompt_qasper}
\end{figure*}

\begin{figure*}[t!]
\begin{tcolorbox}[colback=black!3!white, colframe=black!70!white, title=SciERC, fontupper=\footnotesize, fonttitle=\footnotesize]

      You will be shown an abstract from a computer science research paper. Given this abstract, your task is to extract all unique entities with the following types:\newline

      - "Task": Applications, problems to solve, systems to construct. Examples include "information extraction", "machine reading system", "image segmentation".\newline
      - "Method": : Methods, models, systems to use, or tools, components of a system, frameworks. Examples include "language model", "CORENLP", "POS parser".\newline
      - "Metric": Metrics, measures, or entities that can express quality of a system / method. Examples include "F1", "BLEU", "Precision", "time complexity".\newline
      - "Material": Data, datasets, resources, Corpus, Knowledge base. Examples include "image data", "speech data", "stereo images", "CoNLL", "Wikipedia".\newline
      - "OtherScientificTerm": Phrases that are a scientific terms but do not fall into any of the above classes. Examples include "physical or geometric constraints", "qualitative prior knowledge", "tree kernel", "noise".\newline
      - "Generic": General terms or pronouns that may refer to a entity but are not themselves informative, often used as connection words. Examples include "model", "approach", "them".\newline\newline    
      Please return the output as a JSON object of the format: \{"type1" : ["example\_entity", ...], "type2" : ["example\_entity", ...]\}. The keys should be entity types and values should be lists of extracted entities belonging to the corresponding type. Entity types with no matching entities should be assigned an empty array "[]".\newline
    
      For instance, the output might look like: \{"Task": ["speech recognition", ...], "Method": ["Conditional random field"], "Material": [], ...\}.\newline

      Only output the JSON object and do not include any additional text.\newline

      Abstract:\newline
      
      \{\{ org\_text \}\}\newline

      |||\newline

      \{\{ ner\_dict | tojson \}\}

\end{tcolorbox}
\caption{Canonical template for \texttt{SciERC} task in Figure~\ref{fig:task_list}. See \S\ref{sec:appx_prompt_templates} for description.}
\label{fig:prompt_scierc}
\end{figure*}

\begin{figure*}[t!]
\begin{tcolorbox}[colback=black!3!white, colframe=black!70!white, title=HealthVer, fontupper=\footnotesize, fonttitle=\footnotesize]

      You will be shown a claim about public health and the abstract of a biomedical research paper. Each sentence from the abstract will be on a separate line. Your task is to return a JSON object with two fields:\newline

      - "verdict": The fact-checking verdict. If the information in the abstract supports the claim, write "SUPPORT". If the abstract contradicts the claim, write "CONTRADICT". If the abstract does not provide enough information to arrive at a verdict, write "NEI" (for "not enough information").
      
      - "evidence": An array of sentences providing evidence for the verdict. Please copy all relevant sentences verbatim from the abstract. If the verdict was "NEI", then return an empty array.\newline

      For instance, if the model were given the claim "wearing masks can prevent the spread of COVID", the output might be:\newline

      \{\newline
        "verdict": "SUPPORT",

        "evidence": ["Our findings indicate that mass mask-wearing reduces the transmission rate for COVID-19."]\newline
      \}\newline

      Claim: \{\{ claim \}\}\newline

      Abstract:

      \{\{ abstract\_with\_newlines \}\}\newline

      |||\newline

      \{\{ output\_json\_with\_sentences \}\}
      
\end{tcolorbox}
\caption{Canonical template for \texttt{HealthVer} task in Figure~\ref{fig:task_list}. See \S\ref{sec:appx_prompt_templates} for description.}
\label{fig:prompt_healthver}
\end{figure*}

\begin{figure*}[t!]
\begin{tcolorbox}[colback=black!3!white, colframe=black!70!white, title=DiSCoMaT, fontupper=\footnotesize, fonttitle=\footnotesize]

      \{\{ table\_code\_text \}\}\newline

      You are provided with the table above from a materials science paper. Here are JSON templates for two types of numeric cells: "Other" and "Glass\_Compound\_Amount":\newline

      \{"value": "xx", "type": "Other"\}

      \{"value": "xx", "type": "Glass\_Compound\_Amount", "constituent": "xx", "unit": "xx", "material": "xx"\}\newline

      Please describe all numeric cells in the above table following the JSON templates (proceeding by row in a left-right, top-down direction). For each cell, output one JSON description per line. For any unanswerable attributes in the templates, set their value to the placeholder "xx".\newline

      Cell Description:\newline

      |||\newline
      
      \{\{ json\_records \}\}
      
\end{tcolorbox}
\caption{Canonical template for \texttt{DiSCoMaT} task in Figure~\ref{fig:task_list}. See \S\ref{sec:appx_prompt_templates} for description.}
\label{fig:prompt_discomat}
\end{figure*}

\begin{figure*}[t!]
\begin{tcolorbox}[colback=black!3!white, colframe=black!70!white, title=DataFinder Reco MC, fontupper=\footnotesize, fonttitle=\footnotesize]

      You are provided with a research question, keyphrases about the question, a description of candidate datasets and dataset options. Read the description of popular datasets provided below and select the ones that can be used to validate the following research question. Use your knowledge of machine learning datasets to make the best judgement.\newline
      
      Your response should be formatted as a json array. For instance, for the query "Semi supervised image classification", a sample response might be: ["CIFAR-10", "CIFAR-100"]. Do not include any extra text in the response other than the answer array.\newline

      Query: \{\{ query \}\}\newline

      Keyphrases: \{\{ keyphrase\_query \}\}\newline

      Dataset description:\newline
      \{\{ context \}\}\newline

      Options:- \{\{ options \}\}\newline

      |||\newline

      \{\%- set ans\_list = answer.split(", ") \%\}
      
      \{\{ ans\_list | tojson \}\}
      
\end{tcolorbox}
\caption{Canonical template for \texttt{DataFinder Reco MC} (QA-multiple choice) task in Figure~\ref{fig:task_list}. See \S\ref{sec:appx_prompt_templates} for description.}
\label{fig:prompt_datareco}
\end{figure*}

\section{Information About Use of AI Assistants}

We use OpenAI ChatGPT and Anthropic Claude for grammar checking in manuscript preparation. 

\end{document}